\documentclass[letterpaper]{article}
\pdfoutput=1

\usepackage{arxiv}

\usepackage[utf8]{inputenc} 
\usepackage[T1]{fontenc}    
\usepackage{hyperref}       
\usepackage{url}            
\usepackage{booktabs}       
\usepackage{amsfonts}       
\usepackage{nicefrac}       
\usepackage{microtype}      
\usepackage{amsthm}
\theoremstyle{plain}
\newtheorem{theorem}{Theorem}[section]

\usepackage{pdflscape} 
\usepackage{rotating} 
\usepackage{ltxtable} 
\usepackage{ltablex,array} 
\usepackage{mbenotes} 
\usepackage{multirow}

\theoremstyle{definition}
\newtheorem{definition}[theorem]{Definition}

\theoremstyle{remark}

\usepackage{lineno,hyperref}

\usepackage{mathtools}
\usepackage[inline]{enumitem} 
\usepackage{wrapfig,lipsum,booktabs}
\usepackage{array}
\usepackage{graphicx}
\usepackage{color}
\usepackage{amsmath}
\usepackage{algorithmic,float}
\usepackage[linesnumbered,ruled]{algorithm2e}
\usepackage{comment}
\usepackage{balance}
\usepackage{longtable}
\usepackage{float}

\usepackage{pdflscape}
\usepackage{afterpage}
\usepackage{capt-of}

\usepackage{makecell} 

\usepackage{xcolor,colortbl}
\definecolor{LightCyan}{rgb}{0.85,1,1}

\title{Mining User Behaviour from Smartphone data:\\ a literature review}

\author{
 Valentino Servizi \\
  Department of Management Engineering\\
  Technical University of Denmark\\
  Kgs. Lyngby (DK)\\
  \texttt{valse@dtu.dk} \\
   \AND
   Francisco C. Pereira \\
   Department of Management Engineering\\
   Technical University of Denmark\\
   Kgs. Lyngby (DK)\\
   \And
   Marie K. Anderson\\
   Department of Management Engineering\\
   Technical University of Denmark\\
   Kgs. Lyngby (DK)\\
   \And
   Otto A. Nielsen \\
   Department of Management Engineering\\
   Technical University of Denmark\\
   Kgs. Lyngby (DK)\\
}

\begin{document}
\footnotesize
\setlength{\tabcolsep}{3pt}
\newcolumntype{C}{>{\centering\arraybackslash}X}


\maketitle

\begin{abstract}
To study users' travel behaviour and travel time between origin and destination, researchers employ travel surveys.
Although there is consensus in the field about the potential, after over ten years of research and field experimentation, Smartphone-based travel surveys still did not take off to a large scale.
Here, computer intelligence algorithms take the role that operators have in Traditional Travel Surveys; since we train each algorithm on data, performances rest on the data quality, thus on the ground truth. Inaccurate validations affect negatively: labels, algorithms' training, travel diaries precision, and therefore data validation, within a very critical loop.
Interestingly, boundaries are proven burdensome to push even for Machine Learning methods. To support optimal investment decisions for practitioners, we expose the drivers they should consider when assessing what they need against what they get. This paper highlights and examines the critical aspects of the underlying research and provides some recommendations:
\begin{enumerate*}[label=(\roman*)]
    \item from the device perspective, on the main physical limitations;
    \item from the application perspective, the methodological framework deployed for the automatic generation of travel diaries; 
    \item from the ground truth perspective, the relationship between user interaction, methods, and data.
\end{enumerate*}
\end{abstract}


\section{Introduction}
\label{sec:introduction}

Travel surveys capture an essential aspect of user behaviour. To deliver the best possible user experience through the transport network infrastructure, and the transport system \cite{Gong2014}, such a knowledge base enables development of user behaviour models that support planning, design and policy making.

In Denmark since 1975 the National Travel Survey (TU, \textit{Transportvaneundersoegelsen}) collects data about travel behaviour. The Center for Transport Analytics at the Technical University of Denmark is now running the latest version of the survey. To sustain statistical representativeness regarding the whole Danish population and keep it up to date, TU requires the collection of multiple new interviews every day of the year \cite{Christiansen2018}, totalling an average of 12000 interviews per year since 2010 \cite{Christiansen2016}.


Since the introduction of the first generation of smartphones equipped with Assisted Global Positioning System (AGPS) sensor, the research community has considered Smartphone-based Travel Surveys (SBTS) as a promising platform. The first experiment has been attributed to a Personal Digital Assistant (PDA) in 2005~\cite{Gould2013}, in Japan. In 2007, Samsung announced the smartphone i550, as their first smartphone equipped with AGPS\footnote{\href{https://www.reuters.com/article/us-samsung-phone/samsung-to-launch-its-first-ever-gps-phone-idUSL1664279020071017}{Reuters News. Retrieved from web 23/12/2019.}}. To compare SBTS with Traditional Travel Surveys (TTS) similar to TU, researchers produced a large body of literature since 2004~\cite{Bohte2009, Gould2013, Wahlstrom2017}.

Due to the frantic progress of the technology on which such platforms rest, SBTS offer advantages, limitations, threats, and opportunities, which this paper analyses \cite{Feng2015, Gong2014, Montini2015, Montini2014, Nitsche2014, Shen2014, Wang2017,  Zegras2018, Zhao2015, Zhou2016}. 

A major question arises particularly to practitioners: are these technologies ripe to replace TTS? If not, when are they ready? Despite the broad consensus about the potential acknowledged to the SBTS, the deployment has hardly taken off to a large scale, and some researchers theorise that we should wait longer before SBTS adoption can successfully substitute TTS \cite{Gadzinski2018}.

Besides, privacy concerns on trajectories generated by humans with Geographic Positioning System (GPS) are throwing new challenges to the field~\cite{DeMontjoye2013}, and this problem might be one of the roots related to some difficulties observed when recruiting participants for SBTS. The solutions have to cope with both the need for high-resolution data for the research and the need for user's privacy ~\cite{Seidl2016}.

Contrary to a first sight assessment often proclaimed, the substitution of TTS by SBTS is not trivial since, on one side, we have decades of mature process, tuned to well-defined goals and procedures. On the other, we have sophisticated data mining platforms, many more degrees of freedom and tremendous potential, but also bringing new fragility and complexity in the process \cite{Chen2016, Patire2015, Vij2015}. While in TTS the statistical approach is prominent and person to person interactions for data collection overlap with ground truth collection, in SBTS machine learning plays the primary role, interactions are machine to machine for data collection, and person to machine for ground truth collection \cite{Karlaftis2011, Vij2015}. Often the literature uses the term ``data validation'' for the process of ground truth collection.

By tracking the same user with an extended time horizon \cite{Gong2018, Renso2013}, collecting data passively \cite{Semanjski2017, Yurur2016}, detecting previously unreported short trips, and avoiding stereotypes  
of the daily activity that are often reported by users in traditional surveys~\cite{Thomas2018}, SBTS might facilitate the discovery of inter- and intra- user behaviour variations.

Machine Learning is relevant in SBTS because it automates a substantial amount of laborious and repetitive work, such as detection of stop, mode or activity, which until now have been done by trained operatives in charge of collecting trip information from travellers, employed in TTS. 
While in TTS trained operatives collect data and ground truth at the same time, for SBTS the ground truth can't be distilled by any trained operative and machine learning could become quickly useless if building on errors, thus wrong belief. It is important to discriminate between random or systematic errors in the data, and user validation errors. There are tools and methods available to tackle the former, while for the latter we don't have yet sufficient solutions. The data set's ground truth - which is assumed to be validated either by the trained personnel in charge of collecting the data in TTS or by the user in case of SBTS - has an aleatory and discretionary quality level.
The root-data set used to train machine learning algorithms might be objectively included as one of the overall performance's drivers.
Finally, the design of the experiment is one of the variables on which resulting performance depends. For example, the selection of which trajectories will be included in training, validation and test sets is crucial, and the performance might depend strictly on such a design~\cite{Jiang2017}.

With this survey paper we want to provide a holistic view which captures: 
\begin{enumerate}
    \item The data preparation techniques and the machine learning methods used for mining the user behaviour. Essentially, these are allowing automatic ground truth collection about the user route choices and travel time variations.
    \item The technologies' applications contributing in the remote sensing and data collection. These are extending the domain from sensor fusion within the smartphones' protocol stack (e.g. with AGPS and Smartphone Location Services), towards the Internet of Things.
\end{enumerate}

Having in mind the needs of the subject responsible for the Travel Survey, we want to highlight opportunities as well as limitations both old and new.

Therefore, the present review focuses on providing a qualitative analysis to help answering the following research questions:
\begin{enumerate*}[label=(\roman*)]
    \item What are the main Machine Learning (ML) methods in the field?
    \item What is the ground truth?
    \item What is the ground truth relationship with the ML methods?
    \item Which are the main data sets studied?
    \item What characteristics do they have?
    \item What are the features we can extract from these data sets, and how can we extract them?
    \item What are the challenges for ML in the field of smartphone-based travel surveys?
\end{enumerate*}

To tackle the above questions we proceeded snowballing forward first and then backwards~\cite{Wee2016}. Therefore, we looked to the most cited literature of the field in the first step, while in the second step we looked at the references found within the literature harvested in former step. 
The papers have been selected in order to cover deterministic and machine learning methods based on different typology of data sets. We aimed at data sets collected in various geographical areas. About the models and algorithms, we looked at how they exploited various categories of data sources such as GPS, Inertial Navigation Systems (INS), Geographic Information Systems (GIS), and Internet of Things. Moreover, we paid further attention on the underlying variables, distinguishing between location and person agnostic or specific. For example, 
\begin{enumerate*}[label=(\roman*)]
    \item speed is a location and person agnostic variable;
    \item the distance between a GPS position and a bus station is a location specific variable;
    \item the travel history of a user is a person specific variable.
\end{enumerate*} 

The paper begins providing definitions for the main concepts in Sec.\ref{sec:definitions}.

Sec. \ref{sec:background} provides the background of the TS field of research and application.
%

Sec. \ref{sec:smartphones} provides an overview of the technology enabling the data collection, highlighting potential and constraints.

Sec. \ref{sec:data-preparation} presents the techniques relevant for data preparation and features extraction. In particular, we want to introduce their potential impact on the following assessment about users route choices and their travel time. To detect and compare travel time variations and user/context specific relationships, 
we provide a distillation of ML methods for mining the user behaviour from smartphone data. To generate travel diaries automatically, these methods are targeting why one travels, where on the transport network, using which transport mode. Thereby, we review purpose imputation, map-matching, and mode detection methods.

We conclude by discussing what relevant data sets are, and how their features could contribute to understanding the performance of an SBTS, and the underlying ML methods, 
including the challenges that future research might face to boost advancements in the field (see Sec. \ref{sec:future-directions}).

\section{Definitions}
\label{sec:definitions}
\begin{figure}
    \centering
    \includegraphics[width=1\textwidth]{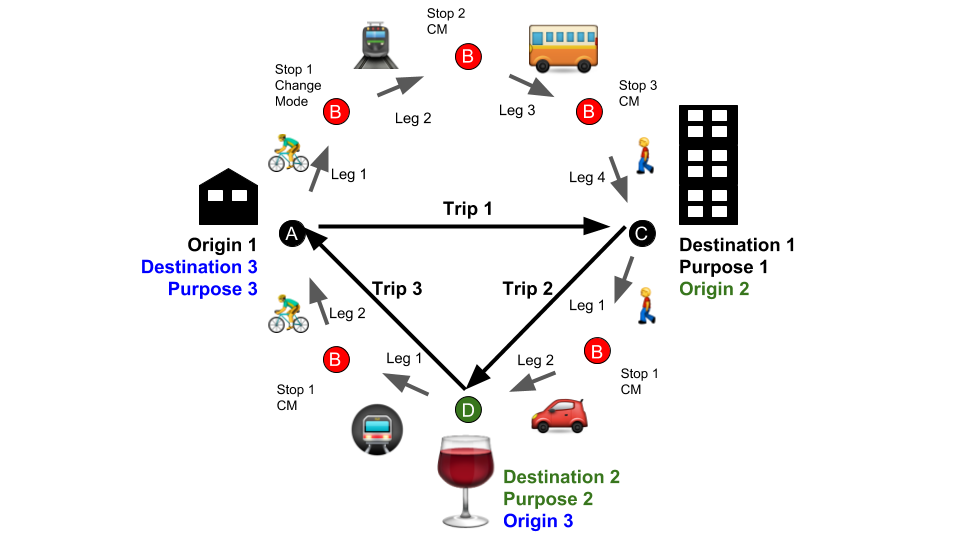}
    \caption{Tour Components.}
    \label{fig:journey}
\end{figure}

This section defines the main terms used.
We pay particular attention to aligning with current 
terminology describing users journeys (see Fig. \ref{fig:journey}).
\begin{definition}
Tour. Aggregation of trips, such that users' travels start and end at the same place\footnote{\href{https://www.merriam-webster.com/dictionary/tour}{Definition of Tour, from Merriam-Webster. Retrieved from web 01/01/2019.}}, e.g. at home~\cite{Christiansen2018}
\end{definition}

\begin{definition}
Trip. Travel entity identified with a set of attributes such as: Start location, Start time, Purpose, Transport Mode, Arrival Time, Arrival Location. A trip could be composed by multiple trip legs~\cite{Christiansen2018}.
\end{definition}

\begin{definition}
Trip Leg. Also identified as trip segment, it identifies, e.g. intermediate short stops for purposes such as: pick-up, drop-off, transfer or change of transportation mode. Each trip segment presents both start and end of both time and location as well as the purpose of the intermediate stop~\cite{Christiansen2018, Semanjski2017}.
\end{definition}

\begin{definition}
Trip Purpose. Often identified with ``activity", the trip purpose represents what triggers the trip from origin to destination. Normally the purpose is related to something to do at the destination and related to, e.g. work, shopping, meeting, picking up, dropping off (somebody or something), eating, education, socialisation, exercise, second home, etc. (see Fig. \ref{fig:journey}, A, B, C, D).
\end{definition}

\begin{definition}
Stop Purpose. It can be reduced to two categories. The first group represents the stops where the purpose is changing transportation mode (see Fig. \ref{fig:journey}, B). The second represents the stops where the purpose is performing the activity which triggered the trip (see Fig. \ref{fig:journey}, A, C, D).
\end{definition}

\begin{definition}
\label{def:mode}
Transport mode. It is referred to a trip leg and it identifies the mode used to get from one end of the segment to the other, e.g. walking, bicycling, car, train, bus, light rail, etc. ~\cite{Christiansen2018, Xiao2016}.
\end{definition}

\begin{definition}
\label{def:mode-chain}
Mode chain type. The literature provides no strict consensus on this term. \cite{Christiansen2018} provides an extensive list including: walk only, bicycle only, driver of passenger car, driver of other vehicle, car passenger, passenger in other vehicle (e.g. taxi, van etc.), aeroplane, other (e.g. ferry, boat, horse), train only (including multiple trains), bus only (including multiple buses), train-bus in combination, train-bus in combination with bicycle, train-bus in combination with car. In consideration of the car sharing business present in the Copenhagen area and spreading across the main capitals worldwide, we might consider train-bus-car in combination with bicycles and scooters.
\end{definition}

\begin{definition}
\label{def:ground truth}
Ground truth. There are several definitions available depending on the application field.
The following two are the most often used in several studies about mode detection, purpose imputation and map-matching: 
\begin{enumerate*}[label=(\roman*)]
        \item \label{list:travelDiaries}travel diaries;
        \item \label{list:promptedRecallSurvey}prompted recall survey~\cite{Auld2009,Chen2016,Feng2016,Lou2009,Miluzzo2008,Nitsche2014,Ohashi2014,Semanjski2017,Shafique2014,Shin2015,Stenneth2011,Vij2015,Xiao2016,Zegras2018,Zhao2015,Zheng2009}.
    \end{enumerate*}
In both cases the information is reported by the users~\cite{YZehngQli2008}, therefore it seems prone to errors. 

\cite{Prelipcean2018} introduces the concept of \textit{``acceptable truth''}. Instead of the theoretical ground truth, we believe that the acceptable truth represents better the above definitions~\ref{list:travelDiaries}, and~\ref{list:promptedRecallSurvey}.

About map-matching, \cite{Biagioni2012}, \cite{Liu2012}, and \cite{Newson2009} refer to a ``hand match'' process correcting visible errors on the output of the map-matching algorithm, or on the map itself. When the map-matching algorithms are focusing on public transportation the ground truth can be extracted from the network: for example \cite{Garg2018} and \cite{Goh2012} focus on bus, and extract the ground truth from both bus stop and intersections networks. In case of synthetic data, the ground truth can be randomly selected between a number of alternatives. For example, \cite{Lou2009} defines the ground truth of a synthetically generated trajectory between an origin and a destination as the random selection within a set of alternative shortest paths. In map-matching applications there are studies using data gathered from an external receiver collecting \textit{``GPS traces with high sampling rate''}~\cite{Aly2015,Huang2014}. 

About mode detection studies, \cite{Shen2014} lists an overview of various ground truth definitions:
    \begin{enumerate*}[label=(\roman*)]
        \item prompted recall survey,
        \item user input in mobile phones,
        \item travel diaries,
        \item experiment (e.g. mode known).
    \end{enumerate*}
About purpose imputation studies, \cite{Shen2014} lists:
    \begin{enumerate*}[label=(\roman*)]
        \item travel diaries,
        \item prompted recall survey.
    \end{enumerate*}
    
\cite{Das2016} refers to the trips reported in-situ by the user participating in the experiment. \cite{Iqbal2014} refers to ``counts'' extracted from video recordings collected from the area relevant for the experiment. Finally, there are several cases where the authors do not mention what the ground truth is~\cite{Shen2014} or they mention scenarios where it is lacking~\cite{Chen2015b,Renso2013,Zegras2018}.
\end{definition}



\begin{definition}
\label{der:travel diary}
Travel Diary. It can focus on "one-day" or on "multiple-days" and it describes the user journeys through attributes, such as:
    \begin{enumerate*}[label=(\roman*)]
        \item Date/s,
        \item Trips of the day,
        \item  Destination of each trip,
        \item Primary Target Destination (see Fig.~\ref{fig:journey}, C),
        \item Purpose (see Fig.~\ref{fig:journey}, C,D),
        \item Trip legs and mode,
        \item Tours, 
        \item Mode Chain Type, 
        \item Trip Start Time, 
        \item Trip End Time.
    \end{enumerate*}
Generally, it is linked to a user and her link-able personal information such as: 
    \begin{enumerate*}[label=(\roman*)]
        \item age, 
        \item occupation, 
        \item education level, 
        \item home address, 
        \item place of occupation address. 
    \end{enumerate*}
Further personal attributes are usually required, e.g. 
    \begin{enumerate*}[label=(\roman*)]
        \item working hours type
        \item and planning;
        \item means of transportation owned,
        \item rather than season tickets 
        \item or car sharing membership;
        \item driver's licence;
        \item handicap, 
        \item house ownership, 
        \item house category, 
        \item personal income, 
        \item household income, 
        \item family type, 
        \item start location of the day, 
        \item primary mode of transport~\cite{Christiansen2018}.
    \end{enumerate*}
\end{definition}
\section{Background}
\label{sec:background}
Human decisions may be determined by different factors, such as
\begin{enumerate*}[label=(\roman*)]
    \item biology, (e.g.) endocrinology, genetics; 
    \item society, (e.g.) culture, gender, religion, wealth; 
    \item mental capacity, (e.g.) IQ or cognition~\cite{Hoseini-Tabatabaei2013}. 
\end{enumerate*} 

\cite{Atallah2009} shows that user behaviour depends on perception, context, environment, prior knowledge, and interaction with others, concluding that human behaviour should be modelled taking into account the context in which users interact. \textit{``These contexts are sometimes referred to as spatial''}~\cite{Hoseini-Tabatabaei2013}, temporal~\cite{Servizi2018}, \textit{`` personal and social aspects, or user context in context aware systems''}~\cite{Hoseini-Tabatabaei2013}. In this extent, TTS are designed to observe the external effects of the user decisions, based on the context in which one interacts. To find correlations between user behaviour and context, a multitude of methods aim at exploiting the data TTS generate. The notion of context here can be extended. In a mode choice model we see variables such as travel time or cost, as part of the context~\cite{Ben-Akiva2017}. 

In a TTS, trained operators are actively interviewing users, interacting with them to collect the ground truth about trips, purpose of each trip, transport mode-chain, route choice, and much more. The operators' job is crucial for the quality of the data collected. Any error in the data collection process could undermine the research based on the collected data.

In Denmark, with a population of 
about 6 millions
, the National Travel Survey (TU) collects $20\,000$ interviews per year in total. The interviews are designed to cover each of the 365 days of the year (temporal representativeness) by focusing on a single day per interview and applying a stratified sampling containing more than 200 strata. The sample represents 2 genders, around 8 age groups between 10 and 84 
(socio-demographic representativeness) and approximately 13 geographical groups (spatial representativeness). The information about the trips contain sources and destinations, purpose, mode (both non-motorised and motorised are considered) and temporal information for each trip~\cite{Christiansen2016}. 

\subsection{Pioneering Smartphone Based Travel Surveys}
Within the last 20 years, TTS methods have been subject to the pressure of disruptive technological evolution. The large penetration of smartphone devices equipped with low cost sensors, the introduction of web 2.0 and its implications~\cite{Anderson2007}, such as Big Data, could identify a tipping point also for this research area, which is constantly pushing towards the full-context awareness~\cite{Hoseini-Tabatabaei2013,Semanjski2017, Yurur2016}.
In this regard, abundant academic literature assesses the impact of new technologies. We highlight the introduction of Global Positioning System (GPS) loggers in the '90s, later the large penetration of sensors rich smartphones, integrated in a high performance communication network (3G, 4G, soon 5G)~\cite{3GPP2015, Bangerter2014a, Kwoczek2015, Warren2014c}.

The reason to complement and/or substitute TTS with smartphone-based technology is: 
\begin{enumerate*}[label=(\roman*)]
\item statistic representativeness, improvable or decreasing in some of the sampled strata~\cite{Nitsche2014};
\item trend of unreported short trips which the user tend to forget or does not want to mention~\cite{Thomas2018};
\item undetected behaviour variations of the same user, due to the design of traditional surveys, which collect a cross-section sample of the population, normally focusing on one single day for each respondent~\cite{Gong2018, Renso2013}.
\label{list:item:focus}
\end{enumerate*}

Only by looking at information on route choice and travel variability described in \ref{list:item:focus}, we can see that detecting behaviour variations within the same user would require the extension of each interview's time scope from $1 \ day$ to $N \ days$ per user. Consequently, each interview would require an amount of time that is proportional to $N$, impacting negatively on both the resources necessary for the task and the percentage of expected rejections, the latter due to the increased burden of such a survey design on each user. Moreover, even in the assumption of this possibility, we should always keep in mind that respondents are likely to fail recalling important details of travel decisions, if they refer back to trips too old to be remembered, and when these decisions deviate from their normal pattern~\cite{Kim2018}.

Most of the SBTS we know offer web and app validation, or only web validation, use machine learning, and are fully automated, as for example FMS/MMM~\cite{Zhao2015}, TRAVELVU/Trivector~\cite{Ek2018}, RMOVE~\cite{Calastri2018}, Itinerum~\cite{Patterson2019,Patterson2016}, MEILI~\cite{Prelipcean2018}, Sydney Travel and Health Survey~\cite{Greaves2015}, Dutch Mobile Mobility Panel~\cite{Thomas2018}. 
The first deployments were FMS in 2012 and Sydney Travel and Health Survey in 2013. 

To the best of our knowledge, all the above SBTS user interfaces have been designed assuming that users would either validate travel diaries accurately generated or correct the errors of the wrong ones. Nonetheless, excluding from the interface any option, and related button, allowing users to report mistakes in the diaries, which, some times, are likely challenging to correct, the risk of including these incorrect data within the ground truth seems unavoidable for the time being. Whether the impact of this possibility is significant or not, we should assess with field research.

\subsection{Mining user behaviour from smartphone data}
Building upon the work of~\cite{Hoseini-Tabatabaei2013}, we highlight the following main progress drivers of SBTS:
\begin{enumerate*}[label=(\roman*)]
\item low cost sensors;
\item support to developers by increasing availability of Software Development Kits (SDKs), Application Program Interfaces API(s), Machine Learning methods and GIS accessibility;
\item introduction of application stores for the distribution of developed applications on a worldwide scale (e.g. Apple App Store and Google Play);
\item Graphic Processing Units (GPU) and cloud processing power.
\end{enumerate*}

The introduction of smartphones in the field of travel surveys, similarly to what has been reported in other fields~\cite{Krumm2008}, shifts the data generation model from the subject running the survey to the participating user. However, in order to collect high quality data while shifting from a Person to Person (P2P) interaction of traditional surveys, to 
SBTS, the user/respondent needs to be supported in different but equally effective ways.

The paradigm rising from the introduction of the technologies described above - where the data mining platform generates automatically the travel diaries from the smartphone data - has lead to machine learning and artificial intelligence having taken over the role of the operator by providing predictions that the user has to validate or eventually correct in order to allow the collection of the ground truth about the recorded trips.


For the subject running the smartphone-based survey, the new process carried out via a dedicated mobile application involves the following steps:
\begin{enumerate*}[label=(\roman*)]
    \item collecting most of the travel data passively,
    \item generating the users' travel diaries automatically,
    \item submitting the travel diaries to the user,
    \item collecting the user validation and/or amendments (ground truth).
\end{enumerate*}
    
For the user participating to the smartphone-based survey, the new process involves the following steps:
\begin{enumerate*}[label=(\roman*)]
    \item installing a survey application on her smartphone and authorising it for data collection,
    \item accessing the application regularly to review, validate and/or amend the travel diaries generated by the survey app.
\end{enumerate*}

The automatic generation of such travel diaries is based on machine learning algorithms, relying on the background data collected to infer: 
\begin{enumerate*}[label=(\roman*)]
\item travel purpose,
\item transportation mode chains,
\item route choices (map-matching).
\end{enumerate*}
These methods in turn rely on other methods targeting:
\begin{enumerate*}[label=(\roman*)]
\item stop detection,
\item trip segmentation.
\end{enumerate*}
Each of the above methods' families can be implemented using multiple machine learning approaches, e.g.
\begin{enumerate*}[label=(\roman*)]
\item Discriminant Analysis (Gaussian), 
\item Bayesian Networks,
\item Hidden Markov Model,
\item Support Vector Machines,
\item Decision Trees,
\item Random Forests,
\item Hierarchical Thresholds,
\item Fuzzy logic, 
\item Neural Networks, 
\item various clustering and classification techniques (e.g. k-means and Key Nearest Neighbour)~\cite{Mazimpaka2016, Hoseini-Tabatabaei2013}.
\end{enumerate*}
Depending on the ML method, better performances can be achieved through, e.g.: 
\begin{enumerate*}[label=(\roman*)]
    \item features extraction techniques~\cite{Wang2019};
    \item or hyper parameters selection~\cite{Balaprakash2019}.
\end{enumerate*}

Iteration after iteration, user after user, the algorithms should increase their accuracy. Higher ML performance enables a higher quality of the ground truth collected from the survey while reducing its burden on the user and thereby facilitating longitudinal data collection, which is targeting behavioural variation within the same user.

Methods able to handle large amounts of location data, with its inherent noise, have huge value~\cite{Nitsche2014,Quddus2007}. 
\textit{``The discovery of certain mobility patterns from the big data offers us an opportunity to identify the links between microscopic individual choices and emergent macroscopic behaviours and to re-examine the decision rules used to model travel related choices.''}~\cite{Chen2016}. A key in this process is the enrichment of such location data sets with contextual information. In SBTS, essential components are user validation, as well as other external data sources (e.g. transport level of service, points of interest, accompanying travellers). A properly validated travel diary data set is thus valuable to investigate traveller behavioural and systemic patterns, such as: 
\begin{enumerate*}[label=(\roman*)]
\item choice models~\cite{Ben-Akiva2017};
\item driving style~\cite{Christopoulos2018};
\item travel time~\cite{Xu2017};
\item roads safety~\cite{Christopoulos2018};
\item traffic congestion~\cite{Ciscal-Terry2016, Cheng2017, Zhang2016};
\item or even drivers emotions~\cite{Katsis2011}.
\end{enumerate*}

\subsection{Impact of interaction design and battery life of smartphones}
Within the scope of smartphone-based travel surveys, there are also other hitherto less obvious but important implications of moving to this new technology, such as the following.
\begin{enumerate*}[label=(\roman*)]
\item User interaction: simplicity and intuitiveness of the interaction design should reduce any potential distraction of the user while interacting with the survey application ~\cite{Nitsche2014}, as distractions could impact the quality of the data collected. Moreover, when the interaction is directed to amend inaccurate predictions of the algorithms involved in the survey, the impact of the interaction design on the quality of the ground truth collected from the respondents is even greater. A poor interaction between the user and the interface of SBTS could trigger a catastrophic loop in which the user validates wrong predictions instead of correcting them. Since the quality of the predictions derives from the quality of the ground truth, in this scenario SBTS would be a failure~\cite{Allstrom2017, Cottrill2013}.
\item Device performance and battery life: any limitation of the phone performance caused by SBTS should be avoided throughout the efficient use of the sensors and data collection background process~\cite{Nitsche2014}.
\end{enumerate*}
\section{Smartphone capabilities and physical limitations for data validation}
\label{sec:smartphones}
\begin{figure}
    \centering
    \includegraphics[width=1\textwidth]{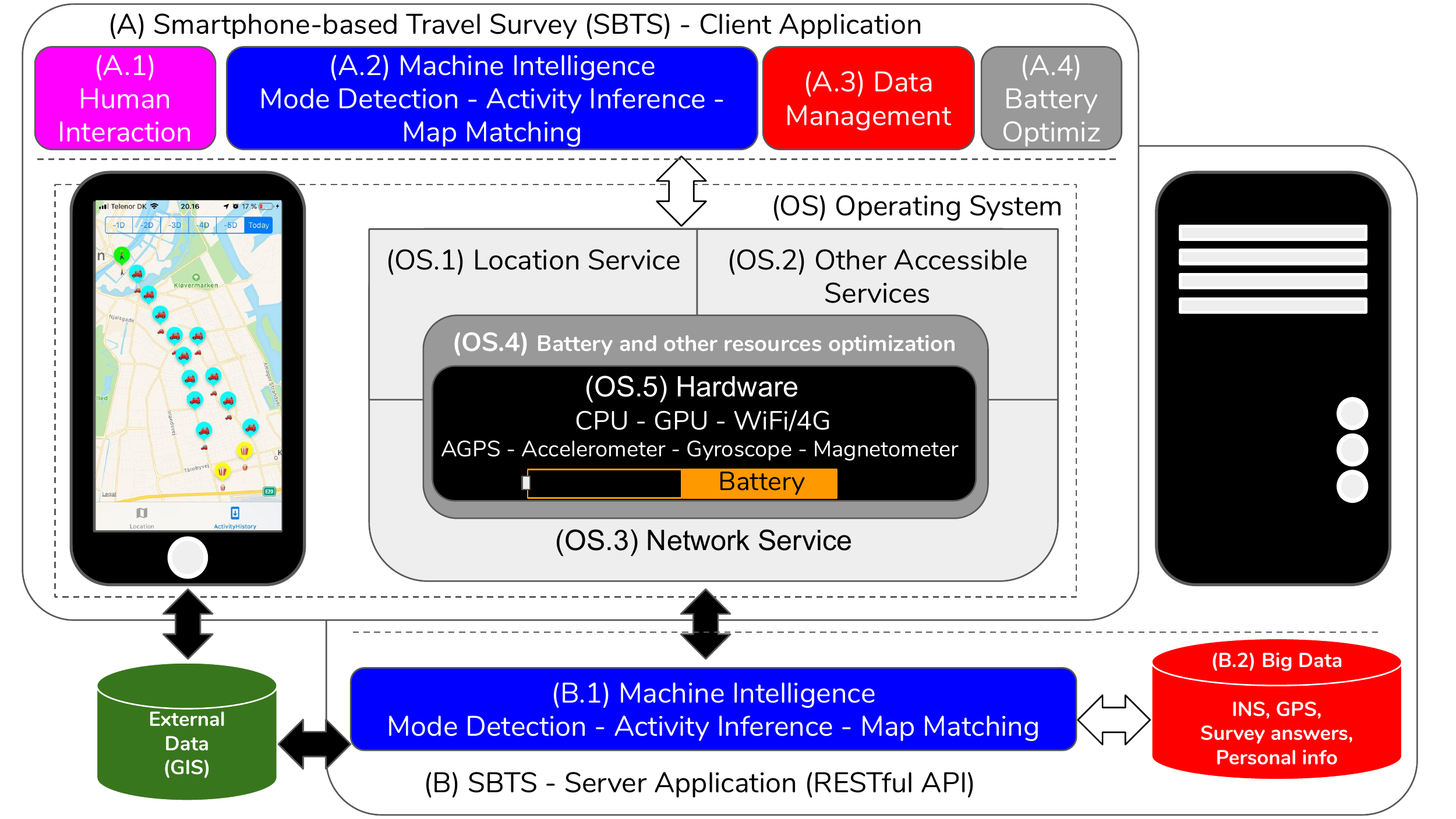}
    \caption{Smartphone Based Travel Survey Platform Architecture.}
    \label{fig:SBTSA}
\end{figure}

From the perspective of smartphone users, the most evident features of any SBTS emerge from the application installed on their device. The user interacts with the smartphone day after day while carrying it around. Thus, any personal perception is constrained by the experience coming from such an interaction.

According to \cite{Assemi2018} the main drivers determining the decision of a user to keep applications on his or her device are 
\begin{enumerate*}[label=(\roman*)]
\item the information conveyed through them,
\item the ease of use,
\item the perceived usefulness,
\item the perceived risks and 
\item \label{list:satisfaction} the general satisfaction of the User Experience (UX\footnote{User Experience acronym attributed to Don Norman~\cite{Cooper2007}}).
\end{enumerate*}
In \ref{list:satisfaction} we include a broad and very relevant field of research which we won’t discuss in this paper. However, we mention that there is consensus about the concerns deriving from the negative impact of smartphone battery consumption on the UX. We observe the same consensus about battery concerns in the literature focusing on the subject of SBTS. In the latter case the negative impact is on the quality of the data collectable with a SBTS. Although this conclusion might look trivial, there is a hierarchy of impacts on smartphones' multi-sided platforms, where independent developers are allowed to distribute native applications via App stores, from the smartphone platform owners, e.g. Apple and Google. For example, any battery optimisation strategy enforced by the Operating System (OS) providers has the priority over any strategy implemented by the application developers allowed on such multi-sided platforms. There is no exception for those who design and develop a SBTS. When it comes to SBTS, which require the protracted use of energy intensive sensors and computer intelligence models, there are situations where the need of high resolution data clashes with the need of battery efficiency enforced by smartphone platform providers~\footnote{\href{https://support.apple.com/en-us/HT208387}{Preventing unexpected shutdowns. Retrieved from web 01/01/2019.}}.

In Fig. \ref{fig:SBTSA} we present the abstraction of a smartphone-based travel survey platform.
The main components are the client side, the App, (A) and the server side, or Back-end, (B) of the application.
The client (see Fig. \ref{fig:SBTSA}, A) is specialised in allowing the human interaction (see Fig. \ref{fig:SBTSA}, A.1). It might include computer intelligence algorithms necessary for the automatic generation of travel diaries (see Fig. \ref{fig:SBTSA}, A.2). The client side can also be specialised in handling the data generated by sensors (e.g. location), computer intelligence models or users involved in the travel diary validation, ensuring persistence, preventing loss of information and maximising privacy by locally processing the data (see Fig. \ref{fig:SBTSA}, A.3). Last but not least, a battery efficiency layer is responsible for tuning and optimising, e.g. data sampling or network I/O operations between client, server or external data sources (e.g. GIS or Digital Maps).
The sensory system of the platform is the smartphone device represented by:
\begin{itemize}
    \item the main relevant hardware components (see Fig. \ref{fig:SBTSA}, OS.5),
    \item  the services exposed by the Operating System (see Fig. \ref{fig:SBTSA}, OS, OS.1-OS.3),
    \item and the operations beyond users and developers influence, such as those focusing on extending the battery life of the device (see Fig. \ref{fig:SBTSA}, OS.4).
\end{itemize}

\begin{figure}
    \centering
    \includegraphics[width=1\textwidth]{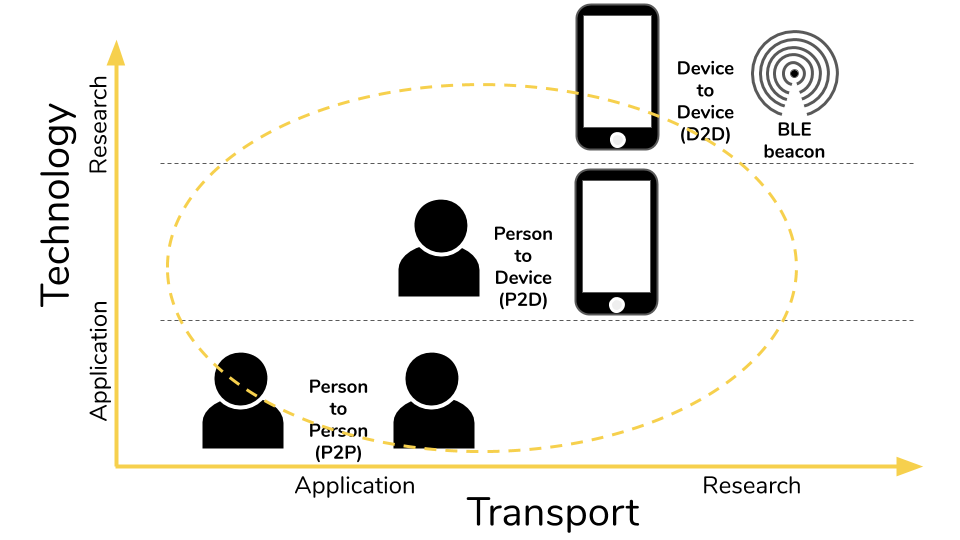}
    \caption{Validation loop evolution.}
    \label{fig:validationLoopEvolutionP2P_D2D}
\end{figure}

In the application field of SBTS the energy consumption derives from the following drivers~\cite{Carroll2010,battery20XX,Wang2019}.
\begin{enumerate}
\item GPU and screen: although it represents one of the most energy hungry components, fortunately it is used only when the user interacts explicitly with the app (e.g., when validating previous observations, when browsing his or her own data, when answering any required survey question). 
\item CPU: it is an energy hungry component and it can receive high load of computations by one or more computer intelligence models, e.g. for mode classification. However, using it properly, we can improve the general energy efficiency of the application. For example, the computation necessary to detect if there are conditions to switch off unnecessary sensors might require an amount of energy below the amount saved by switching off the sensors. One can off-load tasks to the server (e.g. data analysis parts). Of course, off-loading implies transmitting data, which has its own energy cost. 
\item AGPS: Assisted GPS is extensively used in smartphones, and any GPS-capable mobile phone. The difference with standard GPS devices is that while these devices depend exclusively on satellites, to detect the position of the device, AGPS uses also cell tower data. This feature is particularly convenient when GPS signal is weak or disturbed, but it introduces also challenges in the accuracy of the position. There is consensus about the high level of energy consumption of this fundamental sensor. The literature is rich in studies presenting effective strategies to provide the location of a smartphone while reducing the amount of time where the AGPS is active~\cite{Anagnostopoulos2016, Bareth2011, Lin2014, Liu2016, Oshin2012}. For example Apple iOS makes sure that developers can access the location information, but the control of the location updates is constrained by an API which involves the orchestration of multiple sensors and possibly a computer intelligence algorithm~\footnote{\href{https://developer.apple.com/documentation/corelocation/cllocationmanager}{Core location service by Apple Documentation. Retrieved from web 01/01/2019.}}. In this way, accurate location information may require less of GPS and more of WiFi, for example in indoors or high WiFi density areas. Finding the best trade off between location accuracy, resolution and energy consumption is not trivial.
Interestingly, we observe a convergence between approaches developed for the OS to improve the energetic efficiency of smartphones, and for data-mining to fill data gaps coming from missing or highly uncertain GPS observations. Both provide location coordinates, reducing GPS sensor need and leveraging on data from INS, GIS, and telecom-network. Nevertheless, current smartphone OS do not allow access to telecom-network data from independent applications as SBST~\footnote{\href{https://forums.developer.apple.com/message/196395\#196395}{Apple developers support resolution on network signal strength access. Retrieved from web 01/01/2019.}}.
\item Network: it is a fundamental component by design, ensuring the communication between client and server, and enabling offloading strategies for computational intensive operations. Data generated by the application and validated by the user need also offloading on the back end. Fine tuning of the data transfer strategies between front- and back-end is not optional. For example, important energy savings depend on finding optimal thresholds for handling:
\begin{itemize*}[label={}]
 \item network selection (Cellular of WiFi),
 \item data transfer frequency,
 \item status of battery,
 \item or size of the data-transfer.
\end{itemize*}
\item Accelerometer, Gyroscope, Magnetometer: unlike the GPS, these sensors’ raw data is accessible by the developers of the main OS providers. The literature presents multiple interesting strategies aiming at reducing the GPS up-time involving local or remote CPUs besides the raw data generated by these sensors. The energy efficiency improvements reported suggest that such strategies, apparently complex, are less energy hungry than the GPS sensor, even involving multiple hardware components ~\cite{Anagnostopoulos2016,Bareth2011,Oshin2012,Lin2014a,Liu2016,Zhuang2010}. 
\item Beacon and Bluetooth Low Energy (BLE) technology: BLE beacon devices arise from the convergence of Bluetooth and WiFi protocol in the Internet of Things (IoT) context. Unlike the classic Bluetooth protocol, in IoT applications BLE beacons communication is one to many, involves few bits of data to be broadcast frequently, and does not need any pairing operation. These properties are particularly suitable for proximity detection and interaction with Smartphone devices with the purpose of activity sensing~\cite{Jeon2018}. \cite{Pereira2017} experimented BLE interaction in an SBTS. To the best of our knowledge, this is a pioneering work where the Ground Truth has been collected Device to Device (D2D) in this field (see Fig. \ref{fig:validationLoopEvolutionP2P_D2D}). Among the goals, one was the automatic detection of bus trips exploiting beacons installed at the bus stop and inside the buses running the Silver Line in Boston, USA. The authors report the challenge of finding the right signal strength in order to allow the beacons detection by smartphones in conditions where the signal could be attenuated, e.g. by travellers bodies or relative devices' location, while at the same time reducing the risk of interference with other beacons in the range, e.g.,  when passing by a bus stop or bunching with other buses on the way.
\end{enumerate}

\section{Smartphone data mining}
\label{sec:data-preparation}
Due to the disparity of progress drivers, we see a trend of increasing fragmentation, inconsistencies, availability and volume of travel data. In response to this challenge, two main branches seem to arise as the flip sides of the same coin~\cite{Faouzi2011,Kanarachos2018, Kubicka2018,Shen2014}. 
The first focuses towards data-fusion, intended to compose and then mine high dimensional data sets collected from multiple sources, including:
\begin{enumerate*}[label=(\roman*)]
\item GIS, 
\item INS, 
\item GPS. 
\end{enumerate*}
The second targets the development of, e.g., very sophisticated Computer Intelligence Models, Feature Extraction Methodologies, and Optimal Hyper Parameters Selection, which are constantly improving and thus complementing traditional statistical methodologies, often substituting them for specific purposes~\cite{Karlaftis2011}. 

The potential given by smartphones depends on both the high resolution of the data collectable and the large market penetration of the devices. These are two dimensions determining, as already mentioned, data sets quite complex to deal with. The consequence is a multi-step preparation process necessary before we can get any knowledge from the data. 
Typically, smartphone data are affected by several errors. 
For example, the study presented in \cite{Bierlaire2013}, which compare GPS points coming from a dedicated GPS logger with the data generated by a Nokia N95, shows clearly the problem. Since then, however, the situation improved substantially. The resolution and the accuracy achievable by the sensors and the software operating in last generation smartphones is definitively higher. Nevertheless, the raw measurements vary between smartphones and within the same model of smartphone too~\cite{sensors_variability2013}. Achieving consistency of machine learning methods across different smartphones requires a rigorous process of data preparation, cleansing and trajectory segmentation upfront.

\subsection{Data preparation, trip segmentation and stop detection techniques}
\label{subsec:data-preparation}
When approaching a data set generated by a smartphone, the first step is taking into account that any measurement is affected by noise. The noise is not necessarily random, since it may be correlated with weather conditions; building density, materials, and height; crowdedness; smartphone model; software ``bugs''. The data preparation techniques need to be adapted to each type of measurement. For example: the combination of longitude, latitude and time stamps allow the computation of features such as speed, acceleration, and higher-order time derivatives, e.g., jerk, crackle, pop; gyroscope and accelerometer enable the calculation of orientation and acceleration of the smartphone. On top of this laborious \emph{feature engineering} process, we need to do data cleansing. We can assess the quality of the positions and filter out the worst points, e.g., when speed or acceleration is inconsistent with the context. However, there are different degrees of sophistication between rule-based filters (e.g., the threshold on the max speed), statistical filters (e.g., median filter), and model-based filters (e.g., Kalman filter).
Also, the resolution of the measurements is a crucial factor. For example, accelerometer and gyroscope readings from smartphones should be retrieved with a resolution compatible with the motion frequency of human bodies in the daily routines, which is above $20Hz$~\cite{Hoseini-Tabatabaei2013}. Consequently, on the one hand, simple filters can hardly be effective for mining such features. On the other hand, more sophisticated tools as the Kalman filters can be applied effectively, but at the cost of heavy computations.

\begin{figure}
    \centering
    \includegraphics[width=1\textwidth]{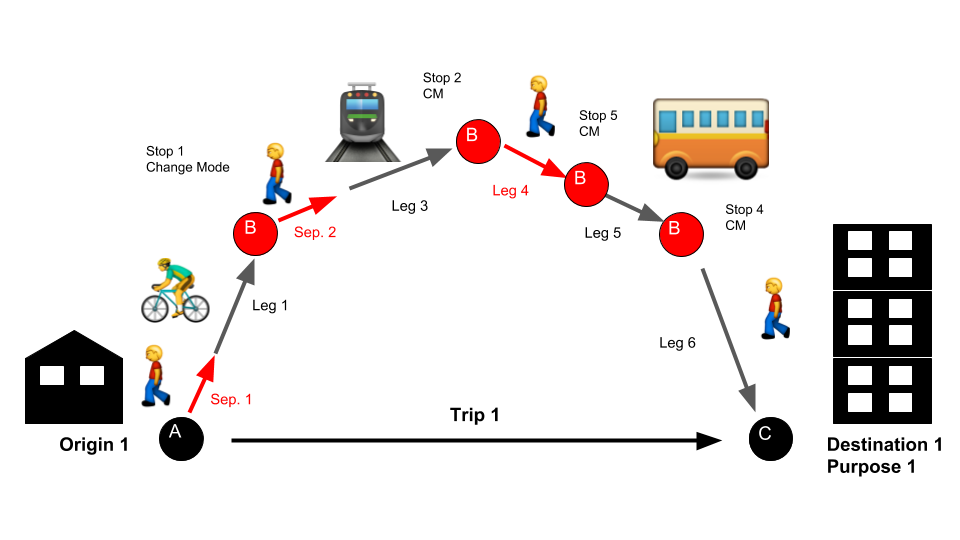}
    \caption{Trip Segmentation using walking segments as separators~\cite{Shin2015}}
    \label{fig:trip segmentation using wolking segments as separators}
\end{figure}

\begin{figure}
    \centering
    \includegraphics[width=1\textwidth]{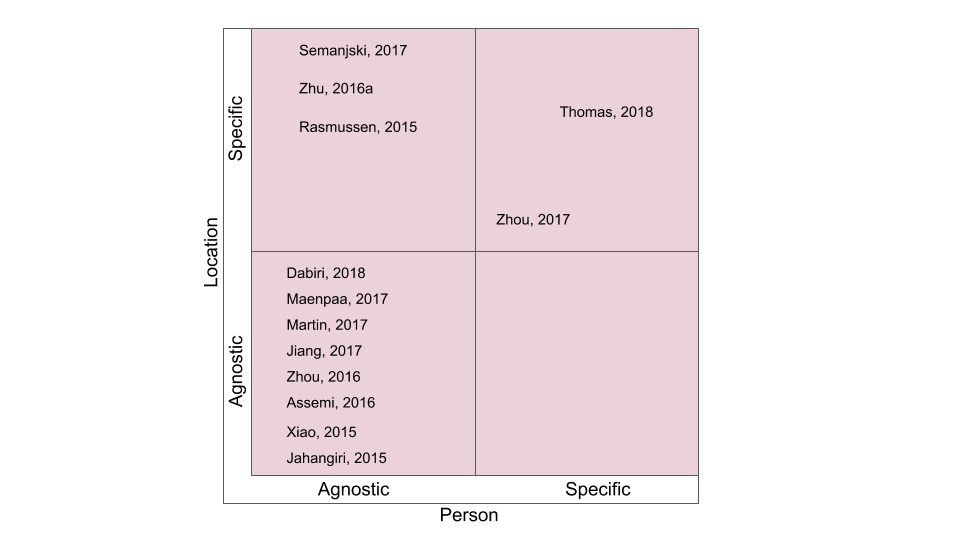}
    \caption{Mode Detection Features Positioning.}
    \label{fig:mode features positioning}
\end{figure}

Once the trajectories are ready for the following classification process, after data cleansing, we can apply ML methods.
We can organise ML methods from the standpoint of their classifying purpose, combined with the underlying features' nature. 
For each classifier, such as for mode detection and purpose imputation, the underlying features can be 
\begin{enumerate*}[label=(\roman*)]
    \item location-agnostic versus location-specific;
    \item and user-agnostic versus user-specific.
\end{enumerate*}
For example, methods relying on user- or location-agnostic features can be trained on any geographic area, 
and then either deployed on a different area to classify the activities of another population, or reused to solve similar problems. The first case depends on the generalisation power of the model, the second case is identified as transfer-learning, which is the discipline dedicated to bringing the knowledge gained in the solution of a specific domain's problem for solving a different problem within another domain. Most of the literature reviewed is working with location- and user-agnostic features (see fig. \ref{fig:mode features positioning}). In contrast, user- and location-specific data seem to enable more accurate classifications, at the cost of volume of information to be handled, poor transfer-ability, and poor generalisation power. This claim is hard to prove due to the heterogeneity of the results available in the relevant literature. Although these results are hardly comparable across related studies, within each relevant study, we find evidence about the positive contribution of user- and location-specific data on the performance of the classifiers~\cite{Kim2018, Semanjski2017, Zhao2015}.

Beyond very different human trajectory classifiers, as those specialised either on mode detection or on purpose imputation, the common ground is human behaviour. People travel with a purpose, often to reach a location where they perform some activity, and their strategy to reach the site depends on the context. Trivially, let us assume that one needs to get to the centre of the city: 
\begin{enumerate*}[label=(\roman*)]
    \item Monday-Friday at 8.00 for work;
    \item Friday at 16.00 for sport activities;
    \item Saturday at 22.00 for social activities.
\end{enumerate*}
It is likely that the strategy to reach the same location will vary according to the purpose and the context, for example:
\begin{enumerate*}[label=(\roman*)]
    \item Public transport or bicycle depending on weather, as the cost is important.
    \item Car, as it might allow room for sport equipment.
    \item Taxi, as it could give freedom of consuming a drink.
\end{enumerate*}

Therefore, the distinction of trajectories reduces to two fundamental classes: 
\begin{enumerate*}[label=(\roman*)]
    \item motion, which is the act of reaching the site of interest, 
    \item and stop, which is the act allowing people to perform the activity of interest in such a place.
\end{enumerate*}
In Sec. \ref{sec:Mode detection} we present how the motion class branches out, and the methods specialising on their classification; in Sec. \ref{sec:Purpose Imputation}, how the stop class branches out, and suitable classification methods.

The approaches and methods about stop-detection and transition-detection described below have also been applied for segmenting GPS trajectories with the purpose of, e.g. mode-detection and/or activity inference.
\cite{Zhao2015} focuses on Future Mobility Sensing (FMS). 
This work highlights the impact of stop-detection performance on the quality of the ground truth collected from users which validate their trips on the mobile device.
The method presented consists in six steps:
\begin{enumerate*}[label=(\roman*)]
    \item trajectory cleansing based on the accuracy provided by the AGPS;
    \item rule based stop detection candidates, where stops are points within 50 meters range and 1 minute time window;
    \item check stop-candidates against user frequent stop location;
    \item rule based merging of the resulting stops applying various range/time thresholds;
    \item detect still mode by applying a learned classifier based on acceleration measures;
    \item remove extra stops after mode detection algorithm.
\end{enumerate*}

\cite{Rasmussen2015} applies a rule based algorithm to detect activity points based on the on-off status of the GPS (because of the GPS units employed in the study), speed/time threshold and range/time threshold. Transition-points are identified by applying a threshold on computed acceleration and speed as well as on the time, based on the assumption that travellers walk to change mode.

\cite{Zhu2016} defines Stay-Points as a geographical area where travellers stay within a range for a certain time. Then, based on these two rules, they apply an \textit{``affinity propagation clustering method''}~\cite{Zhu2016}. Stay-Points belong to a different definition than the transition points used to identify where the travellers change mode in a complex travel mode chain and are identified on a different set of rules based on speed as well as on the assumption that the noisy data typically detected in correspondence of transition points is temporary, while the change in speed are permanent.

\cite{Zhou2017} applies two algorithms in sequence. The first algorithm performs the identification of the trips, which are identified by a rule based algorithm detecting stay points. They eliminate outliers first by performing the Kolmogorov-Smirnov test on a random sample of stay points, in order to verify if these are normally distributed. Then, they apply a three-sigma rule to find and remove outliers. After the cleansing process, they compute the central stay point. Even though GPS follows a bi-variate Raleigh distribution~\cite{Bierlaire2013}, the normal distribution is some times accepted as a suitable approximation.

\cite{Dabiri2018} performs mode classification on the trajectories available in the Geolife  
data set~\cite{Zheng2011}. The authors apply fixed-size segments of 200 points for both seen and unseen trajectories (where 200 is the median of GPS points in all trips composing the data set). Then they concatenate together consecutive segments with the same label. They discard segments with less than 10 GPS points. Finally, the trajectories are processed with a Savitzky-Golay filter for smoothing purpose.

\cite{Jiang2017} tests four segmentation methods: distance-, time-, bearing- and window- based. They highlight that while the last three are statistically equivalent, the first leads to varying sample size within each segment due to the different speeds in complex transport mode-chains. Stop detection is not mentioned explicitly. The work aims directly to transportation mode detection. Thus, transition points might be identified where there is a discontinuity in the mode-chain detected on these segments.

An efficient implementation, recently published on GitHub~\footnote{\href{https://github.com/ulfaslak/infostop}{Infostop by Ulf Aslak. Retrieved from web 26/11/2019.}}, allows stop detection and labelling of stationary events from a GPS trajectory. By building a network that links stationary events identified, as nodes, within a critical space-time range, and clustering this network using two-level Infomap~\footnote{\href{https://www.mapequation.org/publications.html\#Rosvall-Axelsson-Bergstrom-2009-Map-equation}{Infomap is a network clustering algorithm based on the Map equation. Retrieved from web 16/01/2019.}}, the algorithm provides a label for each stop event.

We find an extensive list of methods for trajectory segmentation which are presented from different perspectives~\cite{Prelipcean2016a, Shen2014, VanDijk2018}. 

For example, \cite{Prelipcean2016a} highlights the difficulty of comparing the performances between point- and segment- based methods. Therefore they introduce a penalty system by looking at where these methods make mistakes, and a metric for improving the performance comparison between different segmentation techniques. In particular, with respect to the ground truth, if Precision and Recall identify "hits" and "misses" of the classifier, from such measurements we can't understand how the error depends on over- or under-segmentation of the trajectory. Therefore, since errors in trajectory segmentation propagates to the classification of the trajectories, and classification performance depends on how the segmentation inference aligns with the ground truth, the penalties are proportional to time and space of segments misaligned with the ground truth, in opposition to previous studies where a count of the editing operations was proposed~\cite{Allen1983}. Interestingly, with this metric, point based Trajectory Segmentation Techniques (TST) outperform segment-based TST.

\cite{Shen2014} reviews a large amount of travel-surveys worldwide and when it comes about stop-detection and trip-segmentation they report that rule-based stop detection techniques relying on range, time, speed or acceleration thresholds are the most common. The authors also highlight the challenges about signal loss and signal noise in the detection of short stops.

\cite{VanDijk2018} argues \textit{``that the presence of nearby points in Euclidean space may be indicative of an activity, while the absence of nearby points may be indicative of travel''}~\cite{VanDijk2018}. Therefore, referring to \cite{sss2009}, in order to acquire a local density of points they suggest to deploy a moving window that preserves the relationship with 30 preceding and 30 succeeding points within a $15-m$ range. The range here seems too small compared to GPS error, which is approximately $45-m$~\cite{Zhao2015}. 
\subsection{Human Activity Recognition in Mobility}
\label{sec:machine-learning}
In this section we focus on some of the building blocks necessary for mining user behaviour from smartphone data, in order to support downstream behaviour modelling, as well as infer any general insights that can be useful for a range of applications, from individual services for the user to policy making and planning. On the basis are the fundamental questions in terms of determining the route,  transportation mode, waiting times, and trip purpose.

\begin{figure}
    \centering
    \includegraphics[width=1\textwidth]{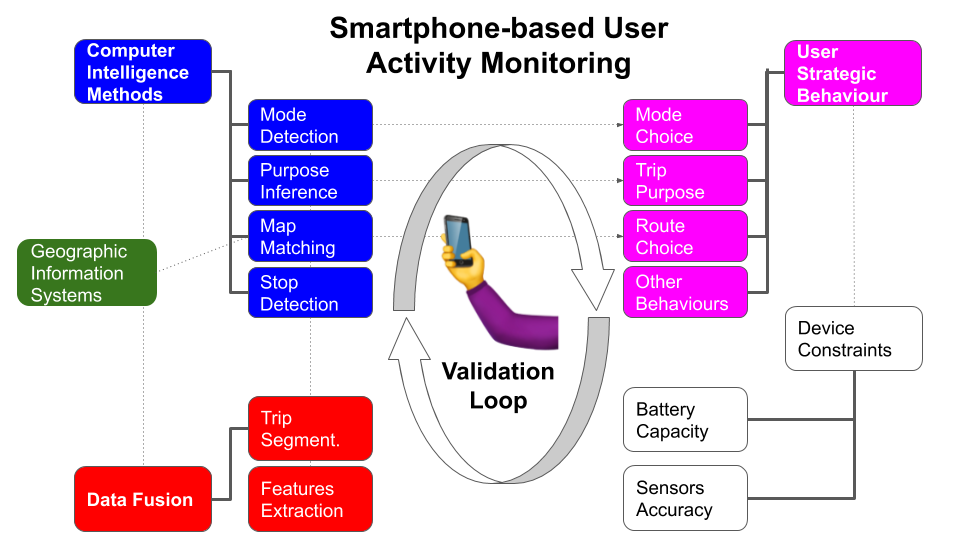}
    \caption{Smartphone-based user activity monitoring.}
    \label{fig:validation loop}
\end{figure}

By leveraging the rich set of sensors installed in the smartphone, we passively collect most of the data related to the user's journey. The most exploited in the papers reviewed are AGPS and accelerometer. The information mapped in the Geographic Information Systems (GIS) can possibly be exploited too. For example GPS locations and accelerometer can be fused on the time dimension and the resulting data set can be fused on the space dimension with the relevant information from the GIS. It is important to note that in the modern smartphones AGPS is not directly accessible, nor the GPS. As mentioned in Sec.\ref{sec:smartphones}, smartphones operating systems provide locations through specific APIs which do not allow direct access to the underlying data sources (e.g. AGPS).
At this point, the combination of feature extraction techniques and computer intelligence algorithms allow capturing the correlation between the features and the user's strategic choices. It is the shared belief in our field, that as technology evolves, the inference of the user strategic choices in the form of a travel diary (see Def. \ref{der:travel diary}) and the user validation by means of such a diary (see Fig. \ref{fig:validation loop}), allow us the generation of the continuous improvement of the acceptable truth (see Def. \ref{def:ground truth}) asymptotically approaching the theoretical ground truth.

To achieve the goal, we rely on a smartphone-based sensing platform (See Fig. \ref{fig:SBTSA}), and we need to take into account the limitations of the smartphone device following the user during her journeys.

Often, computer intelligence algorithms are perceived as black boxes providing an inference as output given some input. For example, the mode detection box should infer the transportation mode, the map-matching box should infer the route choice (See Fig. \ref{fig:validation loop}) and at first glance these seem to be the output of the process. However, these black boxes are tightly coupled with the data necessary to allow and refine the inferences. Given an initial validated data set, their performance can be measured only by comparing the inferences with the data, in other words with the ground truth. 
In Smartphone-based applications, the error propagating from trajectory segmentation, to trajectory classification~\cite{Prelipcean2016a}, and then to the diary generation, could finally propagate to the ground truth. From this standpoint, the output of this process might lead to systematically biased prediction. In SBTS, ML is just a tool to capture the information represented by the data. The quality of the models has a strong correlation with the quality of the ground truth we can collect, through the inferences behind the automatic generation of travel diaries.

There is consensus in the field about the lack of standardisation for validating and comparing the performance of competing classifiers. The work of both \cite{Jiang2017} and \cite{Dabiri2018}, represents an evident example. Even though classifications are performed on the same data-set, their difference in amount and quality of classes predicted, and validation setup, are enough to make F1 scores comparison meaningless (see Tab.~\ref{tab:mode detection}, data set, modes, cross validation).
\cite{Dabiri2018} 
computes F1 scores as the weighted average on a 5-fold cross validation. \cite{Jiang2017} picks a random sample of the users to compose training, validation, and test set, then computes F1 score on the test set only (leave one out method).
To mitigate the problem, in Sec.~\ref{subsec:data-preparation} we find a the penalisation solution proposed in~\cite{Prelipcean2016a} to link F1 score with the distances represented by the classification errors.

Instead, \cite{Wang2019} proposes both data-set and workflow for cross validation, which could provide a standardised baseline. The data-set includes 18 sensors' observations on 3 users, for a period of 2812 hours of labelled data. Labels include the position of the phone as Torso, Bag, Hand, and Hips. The workflow for cross validation includes 3 tasks: User-independent, phone position independent, and time-invariant. At the end of the 3 tasks, each one accomplished with manifold cross validation, as performance driver to measure the predictive power of the model, the paper suggests the standard deviation of F1 score, computed across users, phone positions and time periods. However, most of the data-sets available do not provide the same level of detail, thus do not allow the same validation workflow. For example, the widely used Geolife data-set provides GPS trajectories and transport mode labels only~\cite{Zheng2011}.

\cite{Wang2019} explains that also the feature extraction process should be standardised, and they propose a standard workflow named Minimum Redundancy Maximum Relevance (see Sec.~\ref{sec:Mode detection}).
For classifiers relying on Deep Learning (DL) though, the standard validation method described above is not effective, as the Neural Network extracts the features autonomously. Here, the new challenge is finding optimal hyper parameters for the network, such as architecture configuration, activation functions, batch size, regularisation factor and optimisation step. For example, \cite{Jiang2017}, \cite{Dabiri2018} and \cite{Xiao2018} select such hyper parameters manually, which is a process time consuming and ineffective. \cite{Balaprakash2019} proposes an effective approach to select this hyper parameters automatically, moving a step toward the standardisation of DL-based classifiers optimisation. However, the approach seems not used in this application field yet.

\subsubsection{Mode detection}
\label{sec:Mode detection}

In Tab.\ref{tab:mode detection} we present the summary of the review regarding the transport mode detection methods (see Tab. \ref{tab:mode detection}). These methods aim at inferring the transportation mode chain as defined in Def. \ref{def:mode} and \ref{def:mode-chain}. 
We found focus on the following modes (see Tab. \ref{tab:mode detection}):
\begin{itemize}
    \item Walk, Bike, Electric Bike, Car, Bus, Rail (including both Train and Metro), Motorbike, Boat, Running, Plain, in-Vehicle, Stationary.
    \item In most of the cases we found that the data set Ground Truth come from the validation of the respondents.
\end{itemize}

Training of the methods follows the data cleansing and segmentation described in Sec.\ref{sec:data-preparation}. The research reviewed has been organised according to the person/location agnostic/specific features used to infer the mode of transportation, according to Fig.\ref{fig:mode features positioning}. 

Location and Person Agnostic Features are the following.
\begin{itemize}
    \item Speed: speed, average speed, average speed over a time interval, median speed, n\%-percentile speed, n\%-percentile speed over a time interval, low speed rate (defined as the ratio of points with a speed of less than a threshold), velocity change rate, low velocity rate, high velocity rate, medium velocity rate, max angular velocity, average angular velocity, maximum speed, maximum speed over a time interval, skewness of speed distribution, average change in speed over a time interval, speed variance, speed skewness, speed kurtosis, standard deviation speed.

    \item Acceleration: average absolute acceleration, n\%-percentile acceleration, acceleration change rate, acceleration spectral entropy, acceleration range, maximum acceleration, acceleration variance, average change in acceleration over a time interval, variance change in acceleration over a time interval, acceleration skewness, acceleration kurtosis, Jerk, adjusted acceleration computed by removing the gravity acceleration~\cite{Mizell2003}, applying Fast Fourier Transform to assess the frequency domain (DC), where the DC term is the 0 Hz term and it is equivalent to the average of all the samples in the window, summation of spectral coefficients, energy of the signals.

    \item Distance: travel distance,  share of travel time with the speed within a threshold, ratio of direct distance to travelled distance between origin and destination.

    \item Heading: average heading change, heading change rate, head direction change, bearing rate. 
\end{itemize}

Location Specific Features found are the following.
\begin{itemize}
    \item Distance from motorway, from railway, from bicycle lane, from bus line, from bus stop, from ralways station, from car parking, from bicycle parking; altitude, longitude, latitude, origin, destination.
\end{itemize}

Person Specific Features found are the following.
\begin{itemize}
    \item Departure/arrival time, route, transportation mode~\cite{Zhu2016a}, Personal Trip History~\cite{Thomas2018}
\end{itemize}

We propose the above organisation from an application standpoint. From the General Data Protection Regulation\footnote{\href{https://ec.europa.eu/info/law/law-topic/data-protection_en}{GDPR EU Regulation. Retrieved from web 23/12/2019.}} (GDPR) standpoint all the above information would be person specific, as any observation is collected and linked to the unique user identifier of the subject participating to the SBTS.

The list of computer intelligence methods found is:
\begin{enumerate*}[label=(\roman*)]
    \item Bayesian Network Model (BNM), 
    \item Fuzzy Logic (FL), 
    \item Random Forest (RF), 
    \item bagging model (BM), 
    \item Support Vector Machines (SVM), 
    \item Key Nearest Neighbor (KNN), 
    \item Multinomial Logistic Regression (MNL), 
    \item Multiple Discriminant Analysis (MDA), 
    \item Nested Logit Model (NLM),  
    \item Principal Component Analysis (PCA), 
    \item Bayesian Classifier (BC), 
    \item Neural Network (NN), 
    \item Auto Encoder (AE), 
    \item Deep Neural Network (DNN), 
    \item probabilistic Bayesian mode deduction (PBMD).
\end{enumerate*}

\cite{Koushik2020} provides an extensive description of the main ML methods listed above.

As we already mentioned, every method listed in this section could benefit of a feature extraction, except for DNN, or NN, including AE. 

Maximum dependency minimum redundancy features selection method (aka maximum relevance minimum redundancy), is an effective way of extracting features, but it is also computationally very expensive~\cite{Wang2019} compared to (e.g.) recursive feature elimination.

Rule-based algorithms are simple and effective classifiers, but their limitation is on generalisation power. 

FL success depends on the possibility of applying rules for classification, without losing generalisability. The flaw here is the cost of design, implementation, and mostly any addition required during operations, e.g. the inclusion of a new transportation mode. 

BNM, PBMD and BC are particularly powerful in considering the time dependency of GPS trajectories and time-series. Bayesian methods can be generalised to take into account several time steps, but bringing the risk of over-smoothing, thus of losing their classification power~\cite{Bantis2017}. On the one hand, they could learn online, and avoid retraining when the system would collect new observations; on the other hand, they are computationally very intensive. 

In contrast, SVM, RF, and BM are light and effective, but in the presence of new observations, these methods need to be retrained. About retraining, DNNs have the same drawback; moreover, training is quite complicated. However, here the potential in handling multiple thresholds within the same class and across classes is excellent. DNNs self learn multiple thresholds from the data while extracting relevant features. Consequently, DNNs generalisation power is probably the greatest. Again, the challenge is finding optimal hyper parameters.

Ironically, the ablest and most reliable method of the previous list, depends on at least two elements: the classification task complexity, and the scale of data.

The main sensors leveraged, already mentioned multiple times are the following.
\begin{itemize}
    \item AGPS, accelerometer (Acc), gyroscope (Gyro), magnetometer (Mag). According to~\cite{Wang2019}, Mag contributes in improving Train and Subway detection; Gyro contributes in improving detection of two wheels transportation modes, such as bicycles and motorbikes.
    \item Many studies also rely on Geographic Information Systems, except those  based on Artificial Neural Networks (ANN), both Deep (DNN) and Recurrent (RNN).
\end{itemize}

AGPS represent the cornerstone for the fusion of any sensor with space. For example, Acc, Gyro, and Mag deliver time-series, which theoretically can be fused on the time dimension with AGPS. AGPS provides also a trajectory that can be fused with GIS on the space dimension. 

In the assumption of having relevant information from GIS, compatible with the period of the observations of these sensors, the challenge in handling the resulting data set would derive from, e.g., the position where the user carries the smartphone, the environment that interferes with the AGPS, the battery consumption, the multidimensionality of the resulting data set.

\subsubsection{Purpose Imputation}
\label{sec:Purpose Imputation}
The analysis of purpose imputation methods available in Tab.~\ref{tab:Purpose Imputation}, highlights the use of activity-centred features, cluster-specific features, Location specific features, person and location specific features. Unlike the mode detection methods, this area is heavily populated by person and location specific methods.

About machine learning, the methods encountered include Artificial Neural Network (ANN) and Particle Swarm Optimisation (PSO)~\cite{Xiao2016}, and Clustering (Clu) ~\cite{Montini2014}.

The focus is on the following activities: work, study, shop, social visit, recreation, home, other~\cite{Bohte2009, Feng2015, Kim2018}, service, business meeting~\cite{Feng2015, Montini2014}, paid work, daily shopping, non-daily shopping, help parents/children, voluntary work~\cite{Feng2015}, change mode/transfer, meal/eating break, personal errand/task, medical/dental, entertainment, sports/exercise~\cite{Kim2018}, eating out,  pick up, drop off~\cite{Xiao2016}

About the location-specific features, we found land-use~\cite{Bohte2009, Xiao2016}, residential, administration and public services, commercial and business facilities, industrial, logistics and warehouse, street and transportation, municipal utilities, green space, water bodies, and others~\cite{Xiao2016}

Relevant time indicators taken into account by the models are time of week, time of day, activity duration~\cite{Montini2014,Xiao2016}, start time, end time.

Other features of interests are GPS points density, walk percentage~\cite{Montini2014}

To complete the person specific features we list age, gender, education, working hours, income, mobility ownership~\cite{Montini2014,Xiao2016}.

The features in play seems well represented by the definition provided in Sec. \ref{sec:definitions}. Some are more explicit, but none is unexpected.

In this review we found only very specialised methods, working with data set fusing GPS trajectory with GIS information. 
\subsubsection{Map matching}
\label{sec:Map-matching}
Following the high level classification of map-matching methods presented by \cite{Kubicka2018}, we focus on the subset of outdoor, multi-modal, offline methods for both low and high data sampling rate, which we present in Tab.~\ref{tab:map-matching}.
The large variety of approaches available has been organised in slightly different ways in the main reviews available on the subject.
For example, \cite{Quddus2007} identifies four method categories:
\begin{enumerate*}[label=(\roman*)]
    \item Geometric analysis;
    \item Topological analysis;
    \item Probabilistic algorithms;
    \item Advanced Algorithms.
\end{enumerate*}

\cite{Houston2014} describes also four categories:
\begin{enumerate*}[label=(\roman*)]
    \item Simple;
    \item Weight based;
    \item Advanced;
    \item Multi sensors;
\end{enumerate*}

\cite{Kubicka2018} describes only three categories:  
\begin{enumerate*}[label=(\roman*)]
    \item Geometric;
    \item Multiple hypotheses technique;
    \item Hidden Markov Model and Conditional Random Fields.
\end{enumerate*}

Each of the above reviews takes a methodological stand point, and seems strongly influenced by the latest trend emerging during the year of the review. We don not find any mention of Deep Learning and Artificial Neural Network applications, which are still representing a very exciting niche with applications of both Convolutional and Recurrent Neural Network~\cite{Murphy2017,Wu2017}.

For some map matching methods where are necessary short path generation, \cite{Fu2006} presents an interesting review of heuristics.

For SBTS applications, the above classifications don't communicate some essential aspects though. From the application perspective, the method's categories could provide decision-makers with further meaningful information, such as the following. Is the technique uni-modal or multi-modal? This relates to the ability to perform in either simple or complex transport mode chains. Is it global or incremental? This has implications with operating with or without knowing the destination of the trip. Does the method need the generation of short-paths alternatives to choose the most likely, or can it classify each point directly?  In the first case, the short path generation method is critical; in the second, the point labeling process is also critical. Is the technique rule-based or ML-based? Along this dimension, the distinction between the two ends is indefinite. However, moving from the first to the second category, we notice that heuristics gradually shift from values to distributions; parameters, from input to output of the models.
We have difficulties also in interpreting the methods' performance in doing the job. The problem is manifold. Lack of standardisation translates into different performance' metrics, baselines, and ground truth collection methodologies. Data also plays a crucial role in the performance of the methods. Even in the assumption of adopting the same standard procedure, the performance of the methods still depends on the data set rank, intended as the number of independent features we can extract, and data set size, intended as the amount of independent observations. 
For example, synthetic trajectories (STR) are adopted in several studies. STR involve random selection of short-path generators applied to random origin-destination matrix within a road network, and perturbation with some noise distribution. If on the one hand, this methodology eases the standardisation of the subsequent experiments involving map-matching algorithms, on the other, the performance drop recorded when applying these algorithms to real-life data suggests that the rank is more important than the size of a data set. 
To support decision-makers in assessing different methods, we need to provide them at least with an idea about how different components of the experimental design contribute to the performance estimation.

\begin{figure}
    \centering
    \includegraphics[width=0.75\textwidth]{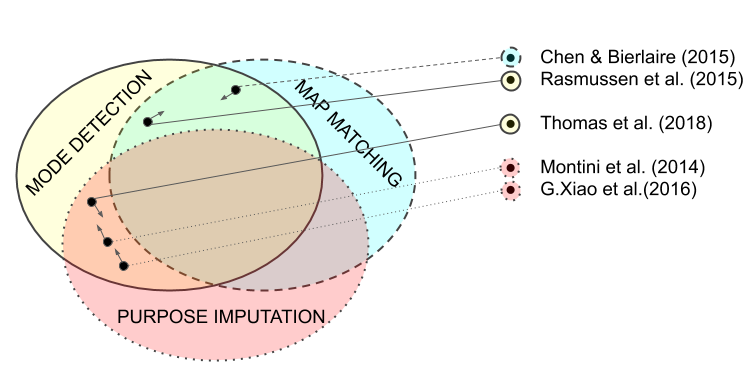}
    \caption{Cross disciplinary studies.}
    \label{fig:gap}
\end{figure}
\section{Conclusion and Future Directions}
\label{sec:future-directions}

To orient Travel Surveys deployment on smartphones, this review provides an analysis of relevant platforms and methods, allowing data collection and data mining. 

We reflect on constraints and challenges deriving from smartphones' limitations. We evaluate the main approaches available, how each method contributes to the generation of the ground truth, how the ground truth affects the methods' performance, and we discuss the interaction models between user and SBTS. A ``person to device'' interaction to validate the data might introduce further errors; what is their magnitude and their impact on the ML methods performance is not clear. The increasing market penetration of iBeacons technology and Internet of Things, together with the positive results reported within the first tests in transport applications, raise expectations on a ``device to device'' ground truth evolution (see Fig.~\ref{fig:validationLoopEvolutionP2P_D2D}). In practice, we could achieve full automation of both travel diary generation and validation, by introducing two new features: the exact location of an IoT device (e.g., an iBeacon), and the strength of the Bluetooth signal received by the smartphone.
Meanwhile, where ML algorithms do not provide correct travel diaries to the user, ``person to device'' interaction could be enhanced by introducing the possibility for the user: 
\begin{enumerate*}[label=(\roman*)]
    \item to trigger a new automatic evaluation of such segments; 
    \item to flag whether he or she was unable to correct the mistakes.
\end{enumerate*}
In the general field, we found consensus about ML performance measurement and experiment design lack of standardisation. Comparison based only on the accuracy level is superficial, as accuracy depends on the underlying data set first, and experiment design second. For example, analysing and testing the code~\footnote{\href{https://github.com/wuhaotju/TrajectoryNet.git}{TrajectoryNet~\cite{Jiang2017} code on Github. Retrieved from web 01/11/2019.}} published by \cite{Jiang2017}, it is evident that a specific split of the data set allows accuracy close to perfection. The authors accomplished the best performance over many other excellent classifiers, trained on the same data set~\cite{Zheng2011}. 
Because of their astonishing accuracy score, they contribute to raising the question of how can we assess whether ML methods and data sets are realistic and applicable to real settings.

To provide a standardised way of comparing different methods, \cite{Prelipcean2018} introduces a penalisation system. \cite{Wang2019} offers an open-source data set, a feature extraction method and a cross-validation algorithm. Even in the assumption that these methods would be enough for comparing algorithms specialised in mode detection, purpose imputation and map-matching would be out of the scope. Moreover, the number of stationary points present in these data sets, which is below $20\%$ of the total, suggests that data cleansing focused on downsampling the stationary class quite heavily. In fact, in a realistic setting, sleeping and working represent at least two-third of a day, where one is unlikely to travel.

Since any standardisation requires intense and coordinated work across the research community and beyond, in this paper, we can only select and summarise all the information relevant, at least, for a qualitative comparison of the methods reviewed. To ease such a comparison, we organise the literature into tables, which include information about the classification objectives,  the data sets employed in the experiments, and both experiment and data validation approach. The classification task is relatively more difficult with a larger amount of classes. The accuracy bias is relatively lower when performing cross validation, and when processing larger data sets. Besides, by listing sensors, features, and data set that each of the related works depends on, we identify the main methods underlying the process of ground truth generation, which in SBTS are trip segmentation, mode detection, purpose imputation, and map-matching. 

SBTS depend on a sophisticated multisided platform, which is subject to often conflicting interests over the resources available, such as the smartphone battery. 

In the current versions, the Operating System orchestrates the applications' use of sensors and battery, and precludes direct access to AGPS; therefore, developers have limited configuration possibilities. Furthermore, the data collected through these platforms is affected by severe errors and noise due to exogenous elements. For example, buildings' elevation, or number of satellites on the line of sight, may negatively affect data sets, thus any method classification performance, user validation and ground truth. 

Nevertheless, we found excellent methods. Someone performs best on low-resolution trajectories, where positions are sampled within large time intervals. Other classifiers are tight (e.g.) to the location where trajectories are combined with data from  GIS, or to personal information of the users' population. Among the best performers in terms of accuracy measurement, in general, we find the following: Support Vector Machines, Fuzzy Logic, Random Forests, and Probabilistic Models (e.g., Hidden Markov Model). Classic rule-based algorithms might not perform at the same accuracy level of the methods just mentioned. However, they are still competitive for applications where execution speed is a priority over the accuracy, and where the application scenario is stable.

We also found interesting studies that are trying either to combine multiple methods (see Fig.~\ref{fig:gap}), or to leverage other methods output. For example, \cite{Chen2015b} uses transport mode to improve the map matching task, while \cite{Rasmussen2015} improves mode detection, by map matching GPS trajectories upfront. Similarly, \cite{Montini2014} and \cite{Xiao2016} uses the transport mode to accomplish purpose imputation. 

In contrast, Methods based on Deep Neural Network (DNN), both convolutional and recurrent, are still in the early stages. For Map-Matching and Purpose Imputation, we found applications with GPS combination with GIS, while for stop and mode detection, we found DNN applications with GPS only. Surprisingly, we did not find examples where the DNN flexibility has been exploited to perform active multi-task learning for classification of mode and purpose at the same time. \cite{Wu2017} performs map-matching with a multi-task recurrent neural network; however, mode detection and purpose imputation are not included. It would be naive to think that using DNN for a multi-task classification could effectively target the three problems at the same time. The available data sets enable already an attempt in such a direction. Nevertheless, especially with map-matching, high quality ground truth collection is one of the biggest challenges for any supervised or semi-supervised method.

\begin{landscape}

\begin{tabularx}{\textwidth}{*{7}{c}}
\caption{Mode Detection. See Sec.~\ref{sec:Mode detection}
}\\
\label{tab:mode detection}
\\
\multicolumn{3}{c}{Study}&\\\hline
\textbf{\makecell{Modes}}&
\textbf{\makecell{Performance\\(F1 score)}}& 
\textbf{\makecell{Cross\\Validation}}&
\textbf{Data-set} & 
\textbf{Method} &
\textbf{\makecell{Main Features}}& 
\textbf{Sensors}\\
\hline
\bigskip
\endhead

\rowcolor{LightCyan}
\multicolumn{3}{l}{\cite{Xiao2015}}&\\\hline
\rowcolor{LightCyan}
\colorbox{LightCyan}{\makecell{5\\Walk,Bike,\\el-Bike,Car,Bus}} & 
$92.74\%$ & 
yes & 
\colorbox{LightCyan}{\makecell{Shanghai\\ 1248 person-days\\(Validated by respondents)}}& 
Bayesian Network & 
\colorbox{LightCyan}{\makecell{Average speed,\\95\% percentile speed,\\Average absolute acceleration,\\Travel distance,\\Average heading change,\\Low-speed-rate\\(as the ratio of points\\with speed $<$ threshold)}} & 
AGPS\\

\multicolumn{3}{l}{\cite{Rasmussen2015}}&\\\hline
\makecell{5\\Walk,Bike,\\Car,Bus,Rail} & 
\makecell{92.4\%} & 
\makecell{n.p.} & 
\makecell{Copenhagen\\ 644 person-days\\(Validated by respondents)}& 
\makecell{Fuzzy Logic} & 
\makecell{95\% percentile acceleration,\\95\% percentile speed,\\Median speed,\\Network segment} & 
\makecell{AGPS + GIS} \\ 

\rowcolor{LightCyan}
\multicolumn{3}{l}{\cite{Zhou2017}}&\\\hline
\rowcolor{LightCyan}
\colorbox{LightCyan}{\makecell{6\\Walk, Bike, Bus,\\Car, Rail, Plain}} & 
$86.5\%$ & 
n.p. & 
\colorbox{LightCyan}{\makecell{Bieijing,\\Geolife\\~\cite{Zheng2011},
\\~4000 person-days\\(Validated\\by respondents),\\ 69 users, 3 years}}& 
Random Forest & 
\colorbox{LightCyan}{\makecell{85\% percentile speed,\\Average speed,\\Median speed,\\Medium velocity rate,\\High velocity rate,\\Low-velocity-rate,\\Travel distance }} & 
AGPS + GIS\\ 

\multicolumn{3}{l}{\cite{Jahangiri2015}}&\\\hline
\makecell{5\\Walk, Bike,\\Bus, Car, Run} & 
\makecell{$95.1\%$} & 
\makecell{yes} & 
\makecell{Bieijing, \\Geolife\\~\cite{Zheng2011},
\\~4000 person-days\\(Validated\\by respondents),\\ 69 users, 3 years}& 
\makecell{Random Forest,\\Bagging Model,\\Support Vector Machines,\\Key Nearest Neighbor,\\Max-Dependency Min-Redundancy} & 
\makecell{Acceleration spectral entropy,\\Acceleration range,\\Max angular velocity,\\Average absolute acceleration,\\Average angular velocity} & 
\makecell{AGPS,\\Accelerometer,\\Gyroscope,\\Rotation Vec.}\\

\pagebreak\\

\rowcolor{LightCyan}
\multicolumn{3}{l}{\cite{Semanjski2017}}&\\\hline
\rowcolor{LightCyan}
\colorbox{LightCyan}{\makecell{5\\Walk,Bike,\\Bus,Car,Rail}} & 
$94\%$ & 
yes & 
\colorbox{LightCyan}{\makecell{Leuven,\\4 mln. GPS obs.\\(Validated\\by respondents) \\ 24900 user-days \\ 8000 users \\ 4 months}} & 
Support Vector Machines & 
\colorbox{LightCyan}{\makecell{Distance From (DF) motorway,\\DF railway,\\DF bicycle lane,\\DF bus stop,\\DF ralways station ,\\DF car parking,\\DF bicycle parking,\\DF bus line}} & 
AGPS + GIS\\

\multicolumn{3}{l}{\cite{Assemi2016}}&\\\hline
\makecell{4\\Walk,Bike,\\Bus,Car} & 
\makecell{$94.7\%$} & 
\makecell{no} & 
\makecell{New-Zeland,\\ 372 person-days \\(Validated\\by respondents),\\ 760k GPS obs. \\ 530h trajectories \\ 76 users \\ 2 Months}& 
\makecell{Nested Logit Model,\\Muiltinomial Logistic Regression,\\Multiple Discriminant Analysis} & 
\makecell{Skewness of speed distribution,\\Share of travel time with 
$2 \le speed (m/s) < 8$,
\\Share of travel time with 
$8 \le speed (m/s) < 15$,
\\Maximum speed,\\ $95\%$ percentile acceleration,
\\Maximum acceleration,
\\Acceleration variance,\\
$\frac{direct \ distance \ origin \to destination}{travelled \ distance \ origin \to destination}$
} & 
\makecell{AGPS}\\ 

\rowcolor{LightCyan}
\multicolumn{3}{l}{\cite{Martin2017}}&\\\hline
\rowcolor{LightCyan}
\colorbox{LightCyan}{\makecell{5\\Walk,Bike,\\Bus,Car,Rail}} & 
$96.8\%$ & 
yes & 
\colorbox{LightCyan}{\makecell{Minnesota,\\ 6 users \\(Validated\\by respondents),\\350k GPS obs.,\\1,7 mln acc. obs.}}& 
\colorbox{LightCyan}{\makecell{Random Forest,\\Key Nearest Neighbor,\\Principal Component Analysis,\\Recursive Feature Elimination}} & 
\colorbox{LightCyan}{\makecell{Average change in acceleration ($\Delta T=120s$),\\80\% percentile speed ($\Delta T=120s$),\\Variance change in acceleration ($\Delta T=120s$),\\Maximum speed ($\Delta T=120s$),\\Average speed ($\Delta T=120s$),\\Average change in speed ($\Delta T=120s$)}} & 
\colorbox{LightCyan}{\makecell{AGPS,\\Accelerometer}}\\

\pagebreak\\

\multicolumn{3}{l}{\cite{Maenpaa2017}}&\\\hline
\makecell{4\\Walk,Bike,\\Bus,Car} & 
\makecell{$90.7\%$} & 
\makecell{yes} & 
\makecell{merging of:\\ 1 week BUS trajectories\footnote{\href{http://wiki.itsfactory.fi/index.php/Journeys\_API}{Journeys API}},\\ Geolife\\~\cite{Zheng2011},
\\ 1000 OSM\footnote{\href{https://www.openstreetmap.org/traces}{Open Street Map Open-source Trajectories} trajectories} \\(Validated\\by respondents)} & 
\makecell{Bayesian Classifier,\\Neural Network,\\Random Forest,\\Auto Encoder} & 
\makecell{Maximum acceleration,\\Maximum speed,\\Minimum acceleration,\\Minimum Speed,\\Average acceleration,\\Average speed,\\Acceleration variance,\\Speed variance,\\Speed skewness,\\Speed kurtosis,\\Acceleration Skewness,\\Acceleration Kurtosis} & 
\makecell{AGPS}\\

\rowcolor{LightCyan}
\multicolumn{3}{l}{\cite{Zhu2016a}}&\\\hline
\rowcolor{LightCyan}
\colorbox{LightCyan}{\makecell{5\\Walk,Bike,\\Bus,Car,Rail}} & 
$93.45\%$ & 
yes & 
\colorbox{LightCyan}{\makecell{Bieijing,\\Geolife\\~\cite{Zheng2011},
\\~4000 person-days\\(Validated\\by respondents),\\ 69 users, 3 years}} & 
\colorbox{LightCyan}{\makecell{Auto Encoder,\\Deep Neural Network}} & 
\colorbox{LightCyan}{\makecell{Average speed,\\Travel distance,\\Average acceleration,\\Head direction change,\\Bus stop closeness,\\Subway line closeness}} & 
AGPS + GIS \\

\multicolumn{3}{l}{\cite{Thomas2018}}&\\\hline
\makecell{5\\Walk,Bike,\\Bus,Car,Rail} & 
\makecell{$82\%$} & 
\makecell{n.p.} & 
\makecell{Netherlands,\\ 600 users,\\ 60k trip-legs\\ 20k trips\\~\cite{Geurs2015},\\ 4 months} & 
\makecell{Bayesian Classifier} & 
\makecell{Personal trip history,\\Speed,\\Altitude,\\Longitude,\\Latitude} & 
AGPS + GIS\\

\pagebreak\\

\multicolumn{3}{l}{\cite{Jiang2017}}&\\\hline
\makecell{4\\Walk,Bike,\\Bus,Car} & 
\makecell{$98\%$} & 
\makecell{no} & 
\makecell{Bieijing,\\Geolife\\~\cite{Zheng2011},
\\~4000 person-days\\(Validated\\by respondents),\\ 69 users, 3 years}& 
\makecell{Recurrent Neural Network,\\Hampel filter} & 
\makecell{Speed,\\Average speed,\\Standard deviation speed} & 
AGPS\\

\rowcolor{LightCyan}
\multicolumn{3}{l}{\cite{Dabiri2018}}&\\\hline
\rowcolor{LightCyan}
\colorbox{LightCyan}{\makecell{5\\Walk,Bike,\\Bus,Car,Rail}} & 
$84.8\%$ & 
yes & 
\colorbox{LightCyan}{\makecell{Bieijing,\\Geolife\\~\cite{Zheng2011},
\\~4000 person-days\\(Validated\\by respondents),\\ 69 users, 3 years}}& 
\colorbox{LightCyan}{\makecell{Convolutional Neural Network,\\Random Forest,\\Key Nearest Neighbor,\\Support Vector Machines,\\Multy Layer Perceptron}} & 
\colorbox{LightCyan}{\makecell{Speed,\\Acceleration,\\Jerk,\\Bearing Rate}} & 
AGPS\\

\multicolumn{3}{l}{\makecell{\cite{Zhou2016}}}&\\\hline
\makecell{5\\Walk,Bike,Run,\\in Vehicle,Stationary} & 
\makecell{$93.8\%$} & 
\makecell{n.p.} & 
\makecell{n.p.}& 
\makecell{Random Forest\\with 3 layers} & 
\makecell{Speed,\\$Acceleration - Gravity$,\\Fast Fourier Transform\\(assess Frequency Domain),\\ Energy of the signals,\\$\Sigma$ spectral coefficients} & 
\makecell{AGPS,\\Acceleration}\\


\end{tabularx}

\label{fn:GIS}
\end{landscape}



\begin{landscape}
\begin{tabularx}{\textwidth}{*{8}{c}}
\caption{Purpose Imputation. See Sec.~\ref{sec:Purpose Imputation} 
}\\

\multicolumn{3}{c}{Study}&\\\hline
\textbf{\makecell{Purposes}}&
\textbf{\makecell{Performance\\(F1 score)}}&
\textbf{\makecell{Cross\\Validation}}&
\textbf{Data-set} &
\textbf{Method} & 
\textbf{\makecell{Main Features}}& 
\textbf{Sensors} & 
\textbf{\makecell{GIS\\Fusion}}\\
\hline
\endhead

\multicolumn{3}{l}{\cite{Bohte2009}}&\\\hline 
\makecell{7\\Work,\\Study,\\Shop,\\Social Visit,\\Recreation,\\Home,\\Other} & 
n.p. & 
n.p. & 
\makecell{Netherlands\\1104 respondents\\ 7395 person-day} & 
Rule-based & 
\makecell{\\Distance GPS $\to$ POIs,\\Distance GPS $\to$ Land Use} & 
AGPS & 
yes \\

\rowcolor{LightCyan}
\multicolumn{3}{l}{\cite{Xiao2016}}&\\\hline 
\rowcolor{LightCyan}
\colorbox{LightCyan}{\makecell{8\\Work,\\Study,\\Shop,\\Social Visit,\\Home,\\Eeating Out,\\ Pick up,\\Drop Off}} & 
$96.53\%$ & 
no & 
\colorbox{LightCyan}{\makecell{China\\321 resp.\\ 2409 person-day\\ 7-12 Days}} & 
\colorbox{LightCyan}{\makecell{Multy Layer Perceptron,\\Particle Swarm Optimisation,\\Multinomial Logit,\\Support Vector Machines,\\Bayesian Network}} & 
\colorbox{LightCyan}{\makecell{Age,\\Gender,\\Education,\\Working Hours,\\Income,\\Time of Week\\Activity Duration,\\Time of Day,\\Transportation Mode,\\Distance GPS $\to$ POIs,\\Distance GPS $\to$ Land Use}} & 
AGPS & 
yes \\ 

\multicolumn{3}{l}{\cite{Montini2014}}&\\\hline 
\makecell{9\\Work,\\Shop,\\Service,\\Recreation,\\Home,\\Pick up,\\Drop off,\\Business Meeting,\\Other} & 
$80\%$ & 
no & 
\makecell{Zurich\\156 resp.\\ 6938 activities\\7 Days} & 
\makecell{Clustering,\\Random Forest} & 
\makecell{Start Time,\\End Time,\\GPS points density,\\Age,\\Education,\\Income,\\Mobility Ownership,\\Activity Duration,\\Walk Percentage} & 
\makecell{AGPS,\\Accelerometer} & 
yes \\ 

\pagebreak\\

\multicolumn{3}{l}{\cite{Feng2015}}&\\\hline 
\makecell{10\\Study,\\Social Visit,\\Recreation,\\Home,\\Service,\\Paid Work,\\Daily Shopping,\\Non-daily Shopping,\\ Help parents/cildren,\\Voluntary work} & 
$75.54\%$ & 
n.p. & 
\makecell{Netherlands\\329 resp.\\ 10545 activities} & 
Random Forest & 
\makecell{Activity Duration,\\Activity Start Time,\\Travel Time to Activity,\\Distance GPS $\to$ POIs} & 
GPS & 
yes \\ 

\rowcolor{LightCyan}
\multicolumn{3}{l}{\cite{Kim2018}}&\\\hline 
\rowcolor{LightCyan}
\colorbox{LightCyan}{\makecell{15\\Work,\\Study,\\Shopping,\\Social Visit,\\Recreation,\\Home,\\Business Meeting,\\Change mode/Transfer,\\Pick up,\\Drop off,\\Meal/Eating break,\\Personal Errand/Task,\\Medical/Dental,\\Entertainment,\\Sport/Exercise}} & 
$96.8\%$ & 
no & 
\colorbox{LightCyan}{\makecell{Singapore\\948 resp.\\ 7856 Days }} & 
\colorbox{LightCyan}{\makecell{Bagging Decision Tree,\\Random Forest}} & 
\colorbox{LightCyan}{\makecell{Activity Probability,\\Distance-based Empirical Probability,\\Activity Transition Probability,\\Activity Duration}} & 
AGPS & 
yes \\ 

\hline

\end{tabularx}

\label{tab:Purpose Imputation}

\end{landscape}


\begin{landscape}
\begin{tabularx}{\textwidth}{*{9}{c}}
\caption{Map-matching. See Sec.~\ref{sec:Map-matching}.
}
\label{tab:map-matching}\\

&\multicolumn{-3}{c}{Study}&&&&&&&\\\hline
\makecell{Category} & 
\makecell{Data Source\\Systems} & 
Method & 
\makecell{Main\\Features}& 
\makecell{Integrity}& 
\makecell{Cross\\ validation}&
\makecell{Performance\\(Accuracy)}&
Data-set &
Location\\
\hline
\endhead

\rowcolor{LightCyan}
&\multicolumn{-3}{l}{\cite{Quddus2006}}&&&&&&&\\\hline 
\rowcolor{LightCyan}
\colorbox{LightCyan}{\makecell{Unimodal,\\Incremental,\\Point-based}} & 
\colorbox{LightCyan}{\makecell{GPS,\\Dead-Reckoning,\\Extended\\Kalaman\\Filter,\\GIS}} & 
Fuzzy Logic & 
\colorbox{LightCyan}{\makecell{Speed,\\Heading\\Error,\\Perpendicular\\Distance,\\Horizontal\\Dilution\\of Precision}} & 
NO & 
NO & 
$99.2\%$ & 
\colorbox{LightCyan}{\makecell{4h trajectory \\1s resolution \\4605 links \\}} & 
\colorbox{LightCyan}{\makecell{London\\sub-urban\\areas}}\\ 

&\multicolumn{-3}{l}{\cite{Li2013}}&&&&&&&\\\hline 
\makecell{Unimodal,\\Incremental,\\Point-based} & 
\makecell{GPS, \\Dead-Reckoning, \\Digital\\Elevation\\Model, \\Extended\\Kalaman\\Filter} & 
\makecell{Rule Based,\\Extended\\Kalaman\\Filter} & 
\makecell{Altitude,\\Longitude,\\Latitude,\\Traffic flow directions,\\Road curvature,\\Grade separation,\\Travel distance,\\Heading} & 
\makecell{YES} & 
\makecell{NO} & 
\makecell{100\% sub-urban\\97\% urban} & 
\makecell{3363 epochs\\2399 epochs\\1s resolution \\}& 
\makecell{Nottingham\\rural\\sub-urban\\Central London}\\ 

\rowcolor{LightCyan}
&\multicolumn{-3}{l}{\cite{Bierlaire2013}}&&&&&&&\\\hline 
\rowcolor{LightCyan}
\colorbox{LightCyan}{\makecell{Unimodal,\\Global,\\Shortest-path}} & 
\colorbox{LightCyan}{\makecell{AGPS,\\GIS}} & 
Probabilistic & 
\colorbox{LightCyan}{\makecell{Timestamp,\\Longitude,\\Latitude,\\Speed,\\Heading,\\Horizontal error Std. Dev.,\\Network error Std. Dev.}} & 
NO & 
\colorbox{LightCyan}{\makecell{Not \\specified}} & 
\colorbox{LightCyan}{\makecell{Not \\specified}} & 
\colorbox{LightCyan}{\makecell{25 GPS traces \\3 users, \\No Ground Truth, \\10s resolution~\footnote{Subset of smartphone-data collected in Switzerland by Nokia Research Center Lausanne, Ecole Polytechnique Fédérale de Lausanne (EPFL), and IDIAP Research Institute, including about 180 participants active during 2 years \cite{Kiukkonen2002}\label{fn:gt}} \\}}& 
\colorbox{LightCyan}{\makecell{Lausanne (CH)\\Urban\\and outskirt\\areas}}\\ 


\pagebreak\\

&\multicolumn{-3}{l}{\cite{Li2014}}&&&&&&&\\\hline 
\makecell{Unimodal,\\Incremental,\\Point-based} & 
\makecell{GPS,\\GIS} & 
\makecell{Neural Network} & 
\makecell{Longitude,\\Latitude,\\Timestamp,\\Heading} & 
\makecell{NO} & 
\makecell{NO} & 
\makecell{87.18\%} & 
\makecell{Training-set: \\8,678 GPS points \\(traces + syntetic from GIS) \\Test-set: \\1,334 GPS points \\(traces only)~\footnote{sub-set of data collected by $12 \cdot 10^3$ taxis, between 1st and 30th November 2012, including $>334 \cdot 10^6$GPS points, 433391 road segments} \\10s resolution}& 
\makecell{Beijing\\urban areas}\\ 

\rowcolor{LightCyan}
&\multicolumn{-3}{l}{\cite{Lou2009}}&&&&&&&\\\hline 
\rowcolor{LightCyan}
\colorbox{LightCyan}{\makecell{Unimodal,\\Global,\\Shortest-path}} & 
\colorbox{LightCyan}{\makecell{GPS,\\GIS}} & 
\colorbox{LightCyan}{\makecell{Mixed Method \\(Topological, \\Geometric, \\Probabilistic)}} & 
\colorbox{LightCyan}{\makecell{Distance $GPS_t \to GPS_{t+1}$,\\Distance $GPS\to Network$,\\Shortest path\\between candidate points\\on Network,\\Average speed}} & 
NO & 
NO & 
\colorbox{LightCyan}{\makecell{\footnote{$A_{N}=\frac{\# \ correctly \ matched \ road \ segments}{\# \ all \ road \ segments \ of \ the \ trajectory}$}$A_{N}>0.81$\\\footnote{$A_{L}=\frac{\Sigma The \ length \ of \ matched \ road \ segments}{The \ length \ of \ the \ trajectory}$}$A_{L}>0.87$}} & 
\colorbox{LightCyan}{\makecell{28 GPS Traces~\footnote{Subset of Geolife~\cite{Zheng2009} Dataset}\\Ground Truth:\\human labels\\}}& 
Beijing\\ 

&\multicolumn{-3}{l}{\cite{Torre2012}}&&&&&&&\\\hline 
    \makecell{Unimodal,\\Incremental,\\Point-based}
& 
    \makecell{GPS,\\GIS}
& 
    \makecell{Hidden\\Markov\\Model,\\Viterbi}
& 
    \makecell{Distance $GPS \to Node$,\\Maximum out-degree\\of the transportation graph}
& 
NO & 
NO & 
    \makecell{Not\\Specified}
& 
    \makecell{128 GPS traces\\$1.85*10^5$ GPS points\\360 km\\1,088 min}
& 
    \makecell{Minneapolis\\(Twin Cities)}
\\ 


\rowcolor{LightCyan}
&\multicolumn{-3}{l}{\cite{Wei2013}}&&&&&&&\\\hline 
\rowcolor{LightCyan}
\colorbox{LightCyan}{\makecell{Unimodal,\\Incremental,\\Shortest-path}} & 
\colorbox{LightCyan}{\makecell{GPS,\\GIS}} & 
\colorbox{LightCyan}{\makecell{Global\\Max-weight,\\Hidden\\Markov\\Model,\\Viterbi}} & 
\colorbox{LightCyan}{\makecell{Fréchet distance,\\Shortest-path}} & 
NO & 
NO & 
98\% & 
\colorbox{LightCyan}{\makecell{14,436 GPS points\footnote{SIGSPATIAL Cup 2012 Dataset}\\19080 GPS points\\1s resolution}}& 
\colorbox{LightCyan}{\makecell{Seattle\\Shanghai}}\\ 



&\multicolumn{-3}{l}{\cite{Chen2015b}}&&&&&&&\\\hline 
\makecell{Multimodal,\\Global,\\Shortest-path} & 
\makecell{GPS,\\GIS,\\Accelerometer,\\Bluetooth Low Energy} & 
\makecell{Probabilistic} & 
\makecell{Transport mode,\\Distance,\\Speed,\\Acceleration} & 
\makecell{NO} & 
\makecell{NO} & 
\makecell{No\\Accuracy} & 
\makecell{No Ground Truth\\10s resolution$^{\ref{fn:gt}}$}& 
\makecell{Lausanne (CH)\\Urban\\and outskirt\\areas}\\ 

\pagebreak\\


\rowcolor{LightCyan}
&\multicolumn{-3}{l}{\cite{Wu2017}}&&&&&&&\\\hline 
\rowcolor{LightCyan}
\colorbox{LightCyan}{\makecell{Unimodal,\\Incremental,\\Point-based}} & 
\colorbox{LightCyan}{\makecell{GPS,\\GIS}} & 
\colorbox{LightCyan}{\makecell{Recurrent\\Neural\\Network,\\Long\\Short\\Term\\Memory}} & 
\colorbox{LightCyan}{\makecell{Trajectory,\\Destination}} & 
NO & 
NO & 
93.58\% & 
\colorbox{LightCyan}{\makecell{442 Taxi\\859,195 Traces\\13,650 Taxi\\3,709,666 Traces\\Ground Truth:\\GPS processed with HMM\\ \cite{Newson2009}}}& 
\colorbox{LightCyan}{\makecell{Porto\\Shanghai}}\\ 


&\multicolumn{-3}{l}{\cite{Newson2009}}&&&&&&&\\\hline 
\makecell{Unimodal,\\Incremental,\\Point-based} & 
\makecell{GPS,\\GIS} & 
\makecell{Hidden\\Markov\\Model,\\Viterbi} & 
\makecell{Distance\\$GPS_t \to GPS_{t+1}$,\\Distance\\$GPS_t \to network$\\(only in range $<$200m)} & 
NO & 
NO & 
\makecell{$>$90\%\\at 30s res,\\100\%\\at 1s res} & 
\makecell{7531 GPS points,\\1s resolution,\\Hand march\\Ground Truth}& 
\makecell{Seattle}\\ 

\rowcolor{LightCyan}
&\multicolumn{-3}{l}{\cite{Hunter2014}}&&&&&&&\\\hline 
\rowcolor{LightCyan}
\colorbox{LightCyan}{\makecell{Unimodal,\\Incremental,\\Shortest-path,\\Supervised,\\Unsupervised}} & 
\colorbox{LightCyan}{\makecell{GPS,\\GIS}} & 
\colorbox{LightCyan}{\makecell{Undirected graph\\Bayesian Network,\\Viterbi}} & 
\colorbox{LightCyan}{\makecell{Path length,\\Distance\\$Point projection \to GPS$,\\Number of signals,\\number of turns,\\Average speed,\\Max/min num. lanes}} & 
NO & 
YES & 
\colorbox{LightCyan}{\makecell{$>$90\%\\at 30s res,\\100\%\\at 1s res}} & 
\colorbox{LightCyan}{\makecell{700,000 GPS points,\\1s resolution,\\High accuracy\\Ground Truth,\\ 560,000 links map}}& 
S.Francisco\\ 


\end{tabularx}
\end{landscape}



\begin{landscape}
\begin{tabularx}{\textwidth}{*{7}{c}}
\caption{Datasets and research 
} \label{tab:dataset and research}\\
\textbf{Fusion} & 
\textbf{Collection} & 
\textbf{Description} & 
\textbf{Ground Truth} & 
\textbf{\makecell{Applications}}& 
\textbf{\makecell{Methods}}& 
\textbf{Reference} \\
\hline
\endhead

\makecell{GPS,\\GIS,\\INS,\\Personal} & 
\makecell{Personal travel survey,\\Shanghai City,\\ smartphone based\\on both Android and iOS.} & 
\makecell{7 days\\352 respondents\\7039 trips} & 
\makecell{User validation}& 
\makecell{Detecting\\trip purposes }&  
\makecell{Artificial\\neural networks \\ and particle swarm\\optimization }& 
\makecell{\cite{Xiao2016}}\\ 
\bigskip

\makecell{GPS,\\GIS,\\INS}& 
\makecell{Two GPS receivers\\integrated with\\an inertial\\navigation system and\\UK’s Integrated Transport\\Network database}& 
\makecell{Nottingham\\(3363 epochs)\\
central London\\(2399 epochs)}& 
\makecell{Devices:\\NovAtel SPAN\\for Nottingham\\ and an iMar for\\central London\\ (centimetre-level \\error performance)}& 
\makecell{Map Matching}& 
\makecell{Extended\\Kalaman filter\\data fusion}& 
\makecell{\cite{Li2013}}\\ 
\bigskip

\makecell{GPS,\\GIS,\\Personal}& 
\makecell{Personal travel survey,\\3 cities in Netherlands,\\data collected carrying\\handheld GPS loggers}& 
\makecell{7 days\\1104 respondents\\15 waves\\1 point each 6 sec.}& 
\makecell{User validation}& 
\makecell{Detecting trip purpose\\and transport mode}& 
\makecell{Data\\collection method}& 
\makecell{\cite{Bohte2009}}\\ 
\bigskip

\makecell{GPS,\\INS,\\Personal,\\LTDS~\footnote{London Travel Demand Survey}}& 
\makecell{London\\Smarphone based on\\both Android and iOS}& 
\makecell{2 users\\1 with crutches\\(7 days)\\1 with wheelchair\\(3 days) }& 
\makecell{User self-labelled states}& 
\makecell{Detecting\\transport mode}& 
\makecell{Hidden\\Markov\\Model (HMM)}& 
\makecell{\cite{Bantis2017}}\\ 
\bigskip

\makecell{GPS,\\INS,\\GIS}& 
\makecell{Chengdu (CHI)\\GPS Data collected\\only from taxi}& 
\makecell{13993 trajectories\\1 month\\24.702 nodes map\\33.532 edges map}& 
\makecell{9 trajectories\\388 km\\14 hours\\435 points}& 
\makecell{Map Matching}& 
\makecell{Information\\Fusion}& 
\makecell{\cite{Hu2017}}\\ 
\pagebreak

\makecell{GPS,\\GIS,\\Personal,\\Weather}& 
\makecell{Personal, longitudinal\\travel survey\\Hakodate, Japan\\smartphone-based\\Android devices provided}& 
\makecell{20 participants\\1 wave winter\\1 wave summer\\1 point each 30 sec.\\9633 trips\\70 features}& 
\makecell{Not specified}& 
\makecell{Detecting\\trip purpose\\and transport mode}& 
\makecell{Random Forests\\and\\Aslan \& Zech’s\\test}& 
\makecell{\cite{Gong2018}}\\ 
\bigskip

\makecell{GPS,\\INS,\\Personal}& 
\makecell{Personal travel survey,\\San Francisco Bay Area\\Smartphone based on\\Android and iOS}& 
\makecell{45 participants\\3381 trips\\3 months}& 
\makecell{User validation}& 
\makecell{Errors impact eval.\\on GPS data-based\\models' inference}& 
\makecell{Monte Carlo\\Experiment}& 
\makecell{\cite{Vij2015}}\\ 
\bigskip

\makecell{INS}& 
\makecell{3 cities, Japan\\Data collected by\\Behavioural Context\\Addressable Loggers}& 
\makecell{46 participants\\3 waves\\3015 trips\\1.107.114 points}& 
\makecell{Paper-based\\travel diaries\\and feedback calls\\to verify mistakes}& 
\makecell{Detecting\\transport mode}& 
\makecell{Support Vector Machine\\Adaptive boosting\\Decision Trees\\Random Forests}& 
\makecell{\cite{Shafique2014}}\\ 
\bigskip

\makecell{GPS,\\INS,\\Personal}& 
\makecell{Personal travel survey,\\Zurich (Switzerland)\\Data collected by\\dedicated GPS devices}& 
\makecell{156 participants\\1 week\\1 point each 1 sec.\\6938 activities}& 
\makecell{User validation of\\automatic generated\\travel diary}& 
\makecell{Detecting\\trip purpose}& 
\makecell{Random Forests}& 
\makecell{\cite{Montini2014}}\\ 
\bigskip

\makecell{GPS,\\INS,\\GIS,\\Personal}& 
\makecell{Personal travel survey\\Copenhagen (DK)\\Data collected by\\dedicated GPS device}& 
\makecell{183 participants\\3-5 days\\6.419.441 points}& 
\makecell{User validation}& 
\makecell{Detecting\\transport mode}& 
\makecell{Fuzzy logic}& 
\makecell{\cite{Rasmussen2015}}\\ 

\pagebreak

\makecell{GPS,\\INS,\\,Travel-cards\\record}& 
\makecell{Personal travel survey\\Singapore\\smartphone-based\\on both Android and iOS}& 
\makecell{1500 participants\\7800 user-validated\\travel days\\}& 
\makecell{8 Participants\\15 phones\\5807 hour}& 
\makecell{Detecting stops}& 
\makecell{Heuristics and\\decision trees}& 
\makecell{\cite{Zhao2015}}\\ 
\bigskip

\makecell{GPS}& 
\makecell{Geolife~\cite{Zheng2009}\\Beijing (CHI)\\GPS loggers and phones
}& 
\makecell{182 users\\3 years\\17.621 trajectories
}& 
\makecell{16 user-labelled\\trajectories}& 
\makecell{Detecting\\transport mode}& 
\makecell{Convolutional\\Neural Networks}& 
\makecell{\cite{Dabiri2018}}\\ 
\bigskip

\makecell{GPS,\\INS,\\Device position}& 
\makecell{Experiment,\\United Kingdom\\(South West)}& 
\makecell{3 participants\\2812 Hours\\17562 Km\\8 modes\\16 sensors\\4 device positions}& 
\makecell{User validation}& 
\makecell{Detecting\\transport mode}& 
\makecell{Decision\\Tree}& 
\makecell{\cite{Wang2019}}\\ 
\bigskip

\end{tabularx}
\end{landscape}

\newpage
\bibliographystyle{unsrt}
\bibliography{main.bib}

\end{document}